\documentclass[a4paper,fleqn]{cas-sc}
\ExplSyntaxOn
\cs_set:Npn \__first_footerline: {}
\ExplSyntaxOff
\usepackage[ruled,lined,linesnumbered]{algorithm2e}
\usepackage[numbers,sort&compress]{natbib}

\newcommand{\secref}[1]{Section~\ref{#1}}
\newcommand{\figref}[1]{Fig.~\ref{#1}}
\usepackage{hyperref}
\hypersetup{
    colorlinks=true,
    linkcolor=blue,
    anchorcolor=blue,
    citecolor=blue
}
\usepackage{color}
\usepackage{enumitem}
\usepackage{amsmath}
\usepackage{booktabs}
\usepackage{adjustbox}
\usepackage{array} 
\usepackage{bm}
\usepackage{CJKutf8} 
\usepackage{makecell}
\usepackage{tabularx}
\graphicspath{{figures/}}  
\usepackage[T1]{fontenc}   
\usepackage{newtxtext}
\usepackage{multirow}
\usepackage{longtable}  
\usepackage{booktabs}  
\usepackage{caption}   
\usepackage{diagbox}
\usepackage[table]{xcolor}
\usepackage[justification=raggedright,singlelinecheck=false]{caption}
\renewcommand{\arraystretch}{1.0}

\newcolumntype{C}[1]{>{\centering\arraybackslash}m{#1}}

\def\tsc#1{\csdef{#1}{\textsc{\lowercase{#1}}\xspace}}
\tsc{WGM}
\tsc{QE}
\tsc{EP}
\tsc{PMS}
\tsc{BEC}
\tsc{DE}
\begin{document}

\let\WriteBookmarks\relax
% \def\floatpagepagefraction{1}
% \def\textpagefraction{.001}

%% Short title
\shorttitle{}    

% \shortauthors{Jian Zhao et~al.}  

\title [mode = title]{DNN Koopman-Based Deviation Compensation for UGV Path Tracking Control on Coupled Slope and Potholed Road}  




 \author[1]{\textcolor{black}{Jian Zhao}}[type=editor,
                         auid=000,bioid=1]
 
 % % Corresponding author indication
 % \cormark[1]
 % 
 % % Footnote of the first author
 % % Email id of the first author
 % \ead{}
 % 
 % % URL of the first author
 
 \credit{Original Draft, Data Curation, Project administration}
 
 \affiliation[1]{organization={National Key Laboratory of Automotive Chassis Integration and Bionics},
     addressline={Jilin University}, 
     city={Changchun 130000},
     % citysep={}, % Uncomment if no comma needed between city and postcode
    % postcode={1043 NX}, 
     country={China}}
 \affiliation[2]{organization={Xiamen King Long United Automotive Industry Co., Ltd.},
     % addressline={XIAMEN KING LONG MOTOR GROUP CO., LTD}, 
     city={Xiamen 361023},
     % citysep={}, % Uncomment if no comma needed between city and postcode
    % postcode={1043 NX}, 
     country={China}}
 \author[1]{\textcolor{black}{Wenbo Zhou}}[style=chinese]
 \credit{Writing - Review, Editing, Formal analysis}
 
\author[1]{\textcolor{black}{Zhicheng Chen}}[style=chinese]
\cormark[1]
\ead{chenzhicheng@jlu.edu.cn}
\cortext[cor1]{Corresponding author}
\credit{Writing - Review, Editing}

 \author[1]{\textcolor{black}{Bing Zhu}}
    \credit{Writing - Review, Editing, Validation}
 %fifth author
   \author[1]{\textcolor{black}{Jiayi Han}}[style=chinese]
   \credit{Writing - Review, Editing}
 %sixth author
  \author[1]{\textcolor{black}{Dongjian Song}}[style=chinese]
 \credit{Investigation}
 %seventh author
 \author[2]{\textcolor{black}{Yinju Lin}}[style=chinese]
   \credit{Investigation}
 %eighth author
 \author[1]{\textcolor{black}{Peixing Zhang}}[style=chinese]
   \credit{Investigation}
 %\author[2]{}%[]
 %
 %% Footnote of the second author
 %% Email id of the second author
 %\ead{}
 %
 %% URL of the second author
 %% Credit authorship
 %% Address/affiliation
 %\affiliation[2]{organization={},
 %%          citysep={}, % Uncomment if no comma needed between city and postcode
 %% Corresponding author text
 %\cortext[1]{Corresponding author}
 %
 %% Footnote text


\begin{abstract}
Unmanned ground vehicles (UGVs) operating in off-road scenarios are confronted with complex terrain disturbances that can substantially degrade path tracking performance. To address this challenge, this paper proposes a deep neural network (DNN) Koopman-based deviation compensation strategy for UGV path tracking control. Firstly, based on the vehicle dynamic function on coupled slope, an adaptive forgetting recursive least squares method with decoupled error terms is designed to estimate tire cornering stiffness. On this basis, a Laguerre model predictive control (LMPC) path tracking control strategy is designed by incorporating Laguerre functions, which can reduce computational resource usage while maintaining reliable tracking performance across different coupled slope scenarios. Then, by integrating Koopman operator theory with DNN, a DNN Koopman (DK) path deviation compensation method is proposed, which significantly improves the path tracking accuracy of UGV under potholed road disturbances. Furthermore, an event-triggered parallel cooperative (EPC) compensation mechanism that couples LMPC with DK is established based on compensation activation criteria and credibility verification. This mechanism improves path tracking accuracy on potholed road while ensuring the feasibility of overall steering command and stability of vehicle after DK compensation. Finally, a hardware-in-the-loop (HiL) experimental platform is constructed for validation. Experimental results demonstrate that the proposed UGV path tracking strategy improves tracking performance by more than 11.5\% across multiple operating conditions.
\end{abstract}


% \begin{highlights}
% \item 
% \item 
% \item 
% \end{highlights}

% Keywords
\begin{keywords}
 Unmanned ground vehicle\sep Path tracking\sep MPC\sep Koopman operator\sep Coupled slope\sep Potholed road
\end{keywords}

\maketitle

% Main text
\section{Introduction}\label{sec:1}
Beyond the structured road scenarios of urban traffic, intelligent driving has also been increasingly applied to unstructured off-road scenarios such as military missions, planetary exploration, and wilderness rescue \cite{1ZhangMeijie1,2ZhangXiaomo2,3An2026}. In these off-road scenarios, unmanned ground vehicles (UGVs), which are the application carriers of intelligent driving technologies, rely on chassis control to accurately track the desired path \cite{4Tan2025}. Unlike the relatively homogeneous pavements in urban scenarios, off-road scenarios encompass diverse terrains such as slopes, rocks, and mud, which involve numerous irregular surface features including side slopes, longitudinal slopes, and potholes \cite{5Wu2026}. These complex terrain features can cause variations in vehicle parameters such as the four-wheel vertical forces, lateral forces, and tire cornering stiffness, thus significantly reducing the driving stability and tracking accuracy of UGV path tracking \cite{6Wang20261}. Therefore, ensuring that UGV can reliably and accurately track the desired path under disturbances induced by complex terrain features remains an important and challenging problem.
\par
By imposing hard constraints on control objectives and solving a finite-horizon optimal control problem, model predictive control (MPC) can effectively overcome the limitations of traditional control methods in handling multi-constraint and multi-objective problems while achieving high-accuracy control. Therefore, it has been widely applied to UGV path tracking control \cite{7Xiong2026,8Wang20262,9Ye2025}. Zhang et al. \cite{10Zhang2025} proposed an MPC-based path tracking strategy incorporating a dynamic control barrier function with a slack variable to improve obstacle avoidance performance of UGV in complex environments, which enables the vehicle to accurately track the desired path in dynamically dense obstacle scenarios. Zhao et al. \cite{11ZhaoJing2025} developed a constrained fractional-order MPC strategy based on a prescribed performance function, ensuring path tracking accuracy and robustness under model mismatch and external disturbances. Similarly, Chen et al. \cite{12Chen2024} introduced a comprehensive feedforward–feedback control framework by integrating an iterative learning algorithm with MPC, enhancing the stability and accuracy of UGV path tracking on rough road. These studies can improve the path tracking accuracy of UGVs on roads with stochastic disturbances. However, on the one hand, the receding horizon optimization and quadratic programming solving process in MPC leads to high computational resource consumption and poor real-time performance. On the other hand, these control strategies neglect the influence of long-wavelength road unevenness, represented by road grade, on UGV path tracking performance. Long-wavelength road unevenness alters the normal load distribution between the front and rear axles and induces additional grade-related forces acting along the slope, making it difficult to continuously guarantee tracking stability and accuracy in off-road scenarios with significant grades \cite{13Liu2025,14Yang2025}. In the design of control strategies, long-wavelength road unevenness is often treated as an additional resistance term in longitudinal dynamics. For example, Ning et al. \cite{15Ning2025} proposed a manifold-based MPC trajectory tracking strategy, improving tracking accuracy and stability of autonomous vehicles under different long-wavelength road unevenness conditions. Cai et al. \cite{16Cai2025} proposed a cooperative adaptive cruise control strategy based on Ito stochastic differential equations and distributed robust H$\infty$ control, which effectively adapts to different long-wavelength road unevenness conditions. The above studies can improve tracking performance under different longitudinal grade driving conditions through longitudinal force dynamics analysis. However, off-road scenarios include a large number of slope cornering conditions. When a UGV steers on a slope, an inevitable coupled slope effect involving both lateral slope and longitudinal slope components arises. This effect will cause rapid variations in tire lateral forces and the vehicle stability boundary, thereby affecting the stability and accuracy of UGV path tracking.
\par
In addition to the coupled slope that reflects long-wavelength road unevenness, UGVs in off-road scenarios also can hardly avoid potholed road that represent short-wavelength road unevenness \cite{17Liu20241}. When traversing potholed road, the tire cornering stiffness of the UGV exhibits nonlinear characteristics due to variations in vertical load. The resulting nonlinear coupling between tire lateral force and longitudinal force causes the vehicle to deviate from the desired path, reducing tracking accuracy and even leading to instability \cite{18Liu20242}. To improve UGV tracking performance on potholed road while reducing the complexity of control strategies, existing model-based studies often equivalently linearize or piecewise approximate complex UGV dynamics using techniques such as exact feedback linearization \cite{19Zhou2025}, fuzzy logic \cite{20Bai2025}, and differential flatness \cite{21Yang2025}, thereby improving UGV tracking performance to some extent. Yong et al. \cite{22Yong2025} proposed a stochastic tube MPC subject to state and control constraints, effectively improving the path tracking accuracy of UGVs under disturbances. Similarly, Lee et al. \cite{23Lee2024} designed a fuzzy path tracking control strategy based on a state observer and a Takagi–Sugeno fuzzy model, enhancing the path tracking accuracy of UGVs on potholed road. However, the tire cornering stiffness and tire forces of the UGV exhibit strong discontinuities under large pothole disturbances. The linearized model inevitably contains unmodeled dynamics, leading to poorer path tracking accuracy and stability. Compared with model-based control strategies that rely on model accuracy, model-free methods such as LSTM \cite{24Li2025}, Actor–Critic \cite{25Bian2026}, and reinforcement learning \cite{26Shi2025} have strong nonlinear approximation capability and have been widely used for modeling and control of nonlinear systems. For example, Xu et al. \cite{27Xu2026} applied neural networks and dynamic iteration methods to develop a model-free iterative learning approach, enhancing the representation capability and control performance for time-varying nonlinear systems. Wang et al. \cite{28Zhao2025} designed an adaptive path tracking control strategy based on a radial basis function neural network and a prescribed performance function, significantly improving the tracking performance of nonlinear vehicle systems under road disturbances. Similarly, to ensure control accuracy when the vehicle exhibits nonlinear characteristics, Deng et al. \cite{29Deng2025} designed a UGV path tracking control strategy that integrates reinforcement learning with a safe traversable corridor. The above studies can substantially improve the adaptability of control strategies to nonlinear systems. However, these model-free control strategies lack interpretability, making it difficult to establish an explicit relationship between model parameters and control strategies. This limitation implies that when vehicle control performance degrades, the control strategy cannot be optimized and adjusted by analyzing the internal mechanisms of the model, which poses new challenges for path tracking control on potholed road.
\par
Fortunately, control strategies that integrate model-based and model-free methods seem to have become an effective avenue for addressing this problem \cite{30Lyu2025,31Chen2024}. Among them, the Koopman operator, which does not rely on specific physical parameters and offers interpretability, has attracted considerable attention. Specifically, Meng et al. \cite{32Meng2025} used the Koopman operator to perform online feedforward error compensation for a fast tool servo with a dual-loop feedback control strategy, improving trajectory tracking accuracy of the servo tool. To ensure UGV path tracking accuracy on pothole surfaces, Wang et al. \cite{33Wang2026} combined the Koopman operator with robust MPC, substantially reducing the path deviation induced by pothole disturbances. Similarly, Mei et al. \cite{34Wu2024} designed an equivalent input disturbance estimator based on a reduced-order model and used a reduced-dimensional Koopman operator to compensate control errors, significantly improving control performance under model mismatch and dynamic disturbances. Current studies have improved the interpretability and accuracy of control strategies by fusing model-free and model-based methods. However, these hybrid control strategies lack consideration of the accuracy and feasibility after control input superposition. After compensating the model-based method with control inputs generated by a model-free method that lacks explicit physical constraints, the resulting overall control input may exceed the limitations of UGV actuators. This will lead to actuator saturation and loss of control, degrade path tracking accuracy, and even trigger system instability.
\par
Motivated by the above discussion, this paper proposes a deep neural network (DNN) Koopman-based deviation compensation strategy for UGV path tracking control strategy considering coupled slope and potholed road. The innovations and contributions of this paper can be summarized as follows:
\begin{enumerate}[label=\arabic*)]
\item  A Laguerre model predictive control (LMPC) based UGV path tracking strategy is developed for coupled slope scenarios. According to the coupled slope dynamic equations, an adaptive forgetting recursive least squares (AFRLS) tire cornering stiffness estimation method with decoupled error terms is designed. On this basis, Laguerre function are used to replace the quadratic programming (QP) solving process in MPC with a fitting and differentiation procedure to generate the baseline steering angle. the proposed method can reduce computational resource consumption,while ensuring path tracking accuracy under different coupled slope conditions. 
\item A DNN Koopman (DK)–based path deviation compensation method is proposed. By using Koopman operator theory and an extended state vector, a nonlinear UGV path tracking system on potholed road is formulated to incorporate the compensatory steering angle. On this basis, a DNN approach with an autoencoder is employed to train the optimal lifting function, thereby obtaining a finite-dimensional approximation matrix of the Koopman operator. This method can significantly enhance the representational capability of the Koopman operator for the nonlinear UGV system on potholed road.
\item An event-triggered parallel cooperative (EPC) compensation mechanism that integrates LMPC and DK is established. A path deviation compensation activation criteria is designed based on the load transfer rate, which characterizes the degree of tracking deviation, and a credibility-driven verification of the compensatory steering angle based on sigmoid function is further incorporated. This mechanism ensures the execution feasibility of the steering angle sequence and the driving stability of the UGV.
\end{enumerate}
\par
The rest of this article is organized as follows. The vehicle path tracking model is formulated in \secref{sec:2}. In \secref{sec:3}, AFRLS tire cornering stiffness estimation method and LMPC path tracking strategy is outlined. \secref{sec:4} illustrate the DK based deviation compensation method and the EPC compensation mechanism. The effectiveness and superiority of proposed control strategy is evaluated by hardware-in-the-loop (HiL) tests in \secref{sec:5} and the conclusions are given in \secref{sec:6}.
\section{Problem Formulation}\label{sec:2}
\subsection{Vehicle dynamic model}
The schematic illustration of a vehicle traveling on a coupled slope is shown in \figref{FIG:1}. The figure includes three coordinate frames: the global coordinate frame $O-XYZ$, the vehicle coordinate frame $o-xyz$, and the vehicle projected coordinate frame $o^{p}-x^{p}y^{p}z^{p}$. The global coordinate frame is fixed to the flat ground, the vehicle coordinate frame is located at the vehicle center of gravity ($CG$), and the origin of the vehicle projected coordinate frame is the projection of the $CG$ onto the $XOY$ plane. Due to the coupled slope effect, the gravitational force of the vehicle in the global frame generates additional force components along the vehicle coordinate frame $o-xyz$, denoted as $F_{x,s}$, $F_{y,s}$, and $F_{z,s}$:
\begin{figure}
	\centering
		\includegraphics[width=0.5\textwidth]{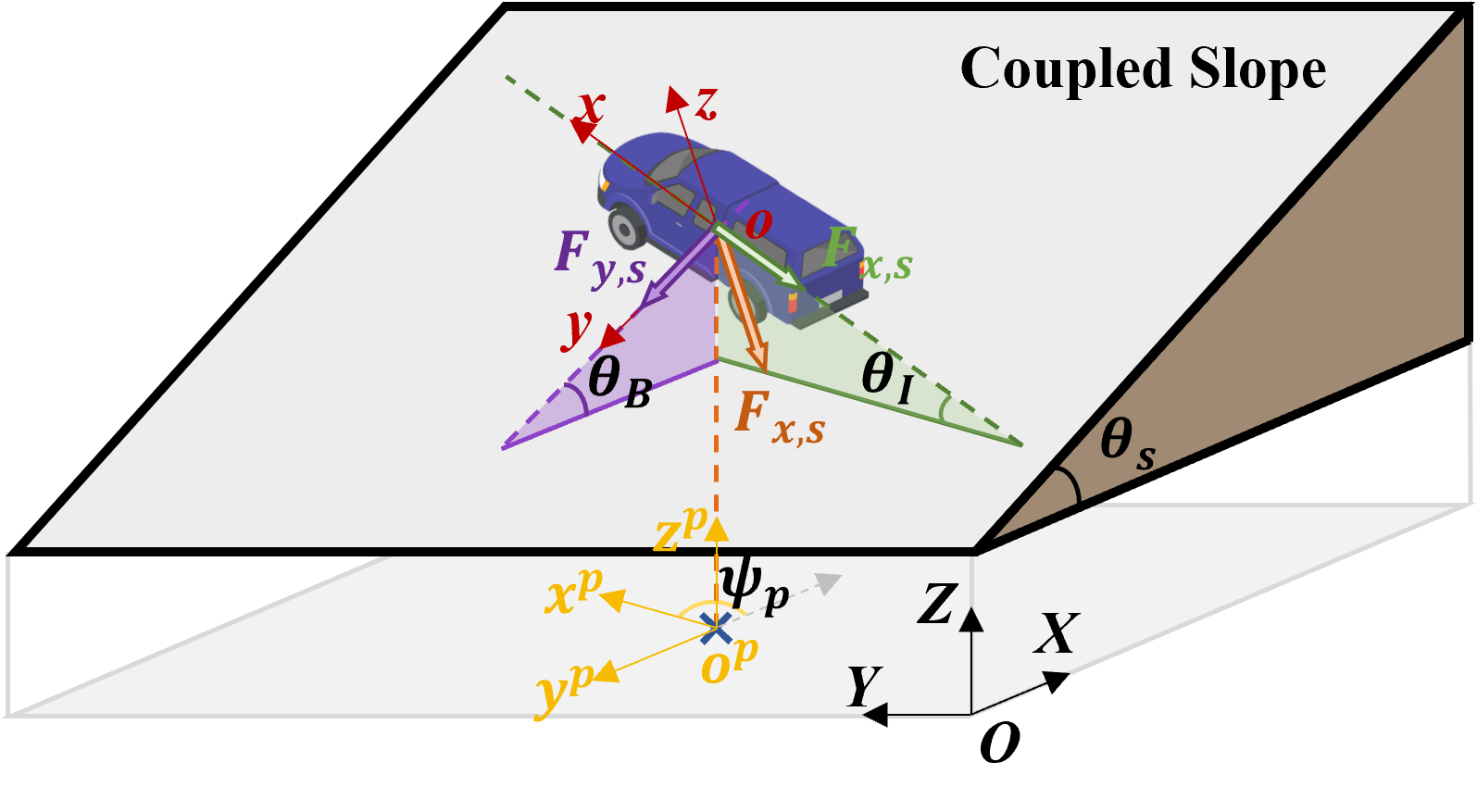}
	\caption{Schematic diagram of UGV driving on coupled slope}
	\label{FIG:1}
\end{figure}
\begin{equation}\label{(1)} 
\begin{bmatrix}F_{x,s}\\F_{y,s}\\F_{z,s}\end{bmatrix}=R_{o^{p}\rightarrow o}R_{O\rightarrow o^{p}}\begin{bmatrix}0\\0\\-mg\end{bmatrix},
\end{equation}
\begin{equation}\label{(2)} 
R_{o^{p}\rightarrow o}=\begin{bmatrix}cos\theta_{I}&sin\theta_{I}sin\theta_{B}&sin\theta_{I}cos\theta_{B}\\0&cos\theta_{B}&-sin\theta_{B}\\sin\theta_{I}&cos\theta_{I}sin\theta_{B}&cos\theta_{I}cos\theta_{B}\end{bmatrix},
\end{equation}
\begin{equation}\label{(3)} 
R_{O \rightarrow o^{p}} = \begin{bmatrix} \cos \psi_{p} & \sin \psi_{p} & 0 \\ -\sin \psi_{p} & \cos \psi_{p} & 0 \\ 0 & 0 & 1 \end{bmatrix},
\end{equation}
where $R_{o^{p}\rightarrow o}$ is the coordinate transformation matrix from the vehicle projected coordinate frame to the vehicle coordinate frame, $R_{O\rightarrow o^{p}}$ is the coordinate transformation matrix from the global coordinate frame to the vehicle projected coordinate frame, $\psi_{p}$ is the projected heading angle of the vehicle, $\theta_{B}$ and $\theta_{I}$ denote the lateral and longitudinal slope angles, respectively.
\par
\begin{figure}
	\centering
		\includegraphics[width=0.5\textwidth]{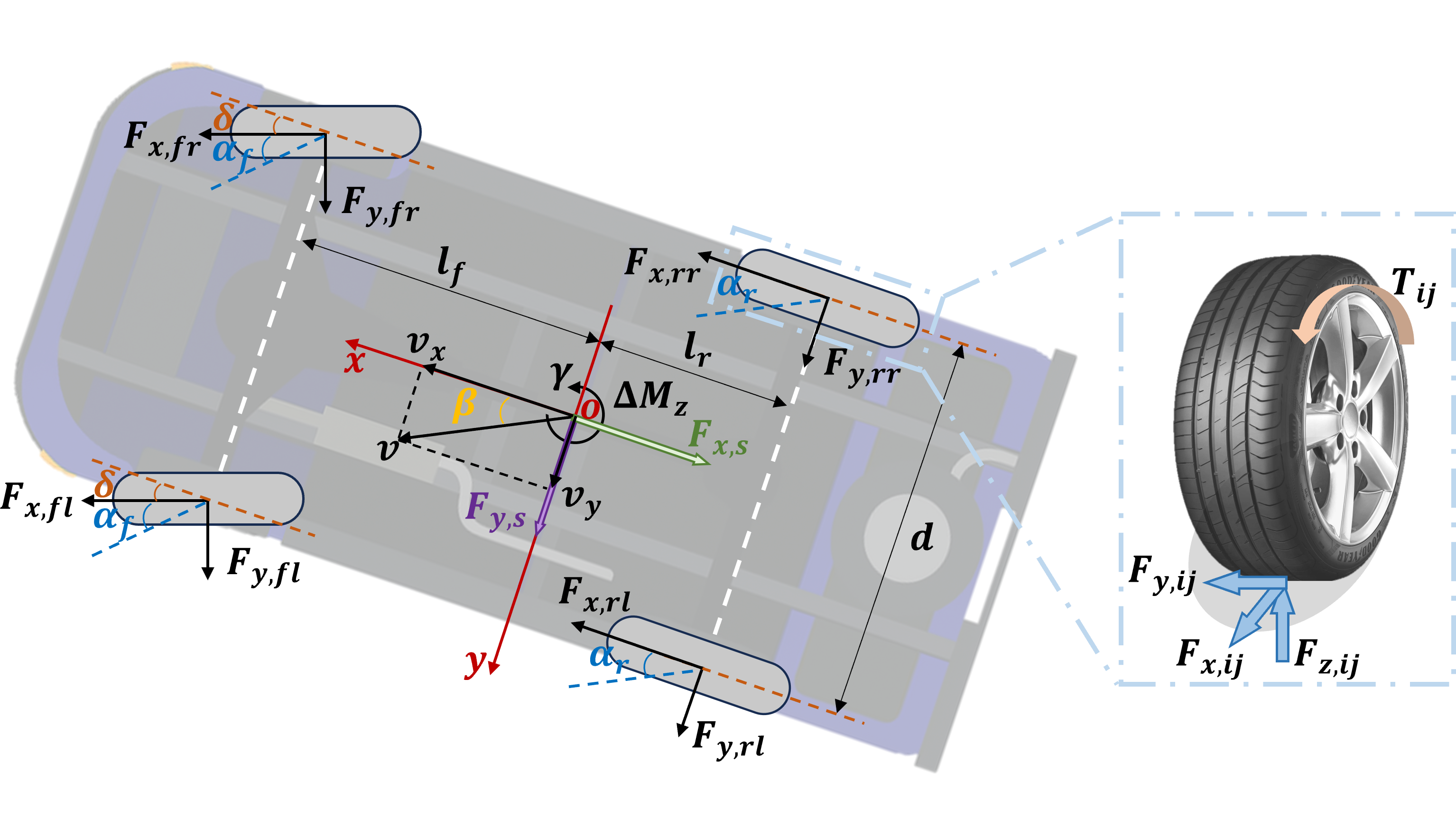}
	\caption{Schematic diagram of vehicle and tire force}
	\label{FIG:2}
\end{figure}
\par
To focus on the key characteristics of vehicle path tracking, a vehicle dynamic model on the coupled slope is established, as shown in \figref{FIG:2}, in which the vehicle’s roll, pitch, and vertical motions are neglected:
\begin{equation}\label{(4)} 
\begin{aligned}
m v_{x}\gamma + m \dot{v}_{y} ={}
(F_{y,fl}+F_{y,fr})\cos\delta
+ F_{y,rl}+F_{y,rr} + F_{y,s}+ (F_{x,fl}+F_{x,fr})\sin\delta ,
\end{aligned}
\end{equation}
\begin{equation}\label{(5)} 
\begin{aligned}
I_{zz}\dot{\gamma} ={}&
\left( l_{f} \sin\delta - \frac{d}{2}\cos\delta \right) F_{x,fl}
+ \left( l_{f} \sin\delta + \frac{d}{2}\cos\delta \right) F_{x,fr} + \frac{d}{2}\left( F_{x,rr} - F_{x,rl} \right)
+ \left( l_{f} \cos\delta + \frac{d}{2}\sin\delta \right) F_{y,fl} \\
&+ \left( l_{f} \cos\delta - \frac{d}{2}\sin\delta \right) F_{y,fr}
- l_{r}\left( F_{y,rl} + F_{y,rr} \right).
\end{aligned}
\end{equation}
where $m$ denotes the vehicle mass, $I_{zz}$ denotes the yaw moment of inertia, $\beta$ represents the sideslip angle, $\gamma$ is the yaw rate, and $v_x$ and $v_ y$ denote the longitudinal and lateral velocities, respectively. $F_{x,ij}$ and $F_{y,ij}$ represent the longitudinal and lateral tire forces of the four wheels, respectively. $\delta$ is the front wheel steering angle, $d$ is the vehicle track width, $l_f$ and $l_r$ denote the distances from the $CG$ to the front and rear axles, respectively.
\par
Since UGVs in off-road scenarios typically operate at low speeds and with small steering angles, the small-angle approximation is adopted (i.e., $sin\delta\approx0$ and $cos\delta\approx1$). Moreover, the longitudinal acceleration is assumed negligible over each sampling interval. Accordingly, \eqref{(4)} and \eqref{(5)} can be rewritten as:
\begin{equation}\label{(6)}
m v_{x}(\dot{\beta} + \gamma) = F_{y,s} + \sum F_{y,i j},
\end{equation}
\begin{equation}\label{(7)}
I_{zz} \dot{\gamma} = l_{f} (F_{y,fl} + F_{y,fr}) - l_{r} (F_{y,rl} + F_{y,rr}) + M_{z,{csl}} + M_{z}
\end{equation}
where $M_{z,csl}=\frac{d}{2}\mu(F_{z,r}-F_{z,l})$ is the vehicle yaw moment induced by the coupled slope, and $M_{z}$ is the additional vehicle yaw moment caused by wheel torque distribution.
\par
Under the effects of coupled slope and lateral acceleration, the vertical tire forces $F_{z,j}$ on the two sides can be calculated as:
\begin{equation}\label{(8)}
F_{z,l} = \frac{2m}{L} \left( \frac{gl_r \cos{\theta_s}}{2} - \frac{v_x \dot{\beta} l_r h_g}{d} \right),
\end{equation}
\begin{equation}\label{(9)}
F_{z,j} = \frac{2m}{L} \left( \frac{gl_f \cos{\theta_s}}{2} + \frac{v_x \dot{\beta} l_f h_g}{d} \right),
\end{equation}
where $h_g$ is the height of $CG$ above the ground.
\par
The tire cornering stiffness is affected by the tire slip angle $\alpha$, the road adhesion coefficient $\mu$, the vertical tire force $F_{z}$, thus exhibits parameter uncertainty during driving. Therefore, a time-varying stiffness correction coefficient $\tau_{ij}=f(\alpha_{ij},\mu_{ij},F_{z,ij})$ is introduced to accurately represent the tire cornering stiffness $C_{ij}$ of each wheel. Accordingly, the lateral tire forces $F_{y,ij}$ acting on the front and rear axles can be expressed as follows:
\begin{equation}\label{(10)}
F_{y,i,j} = (C_{i0} + \tau_{ij} C_{iv}) \alpha_i\quad
i=f, r\quad j = l,r
\end{equation}
\begin{equation}\label{(11)}
\alpha_{f} = \delta - \beta - \frac{l_{f} \gamma}{v_{x}},
\end{equation}
\begin{equation}\label{(12)}
\alpha_{r} = \frac{l_{r} \gamma}{v_{x}} - \beta,
\end{equation}
where $\alpha_{i}$ is the tire slip angle, $C_{i0}$ is the nominal tire cornering stiffness of the wheel, and $C_{iv}$ denotes the variation range of the tire cornering stiffness.
\par
Based on the above formulations, the vehicle dynamic model considering the coupled slope can be expressed as follows:
\begin{equation}\label{(13)}
\begin{aligned}
\dot{\beta} ={} - \frac{2(C_f + C_r)}{mv_x} \beta + \left[ \frac{2(l_r C_r - l_f C_f)}{mv_x^2} - 1 \right] \gamma + \frac{2C_f}{mv_x} \delta + \frac{g \sin{\theta_B}}{v_x},
\end{aligned}
\end{equation}
\begin{equation}\label{(14)}
\begin{aligned}
\dot{\gamma} ={} \frac{2(l_r C_r - l_f C_f)}{I_{zz}} \beta - \frac{2(l_f^2 C_f + l_r^2 C_r)}{I_{zz} v_x} \gamma + \frac{2 l_f C_f}{I_{zz}} \delta + \frac{d \mu m g \cos \theta_s (l_f - l_r)}{2 I_{zz} L} + \frac{M_z}{I_{zz}},
\end{aligned}
\end{equation}
\subsection{Path tracking kinematics model}
\begin{figure}
	\centering
		\includegraphics[width=0.4\textwidth]{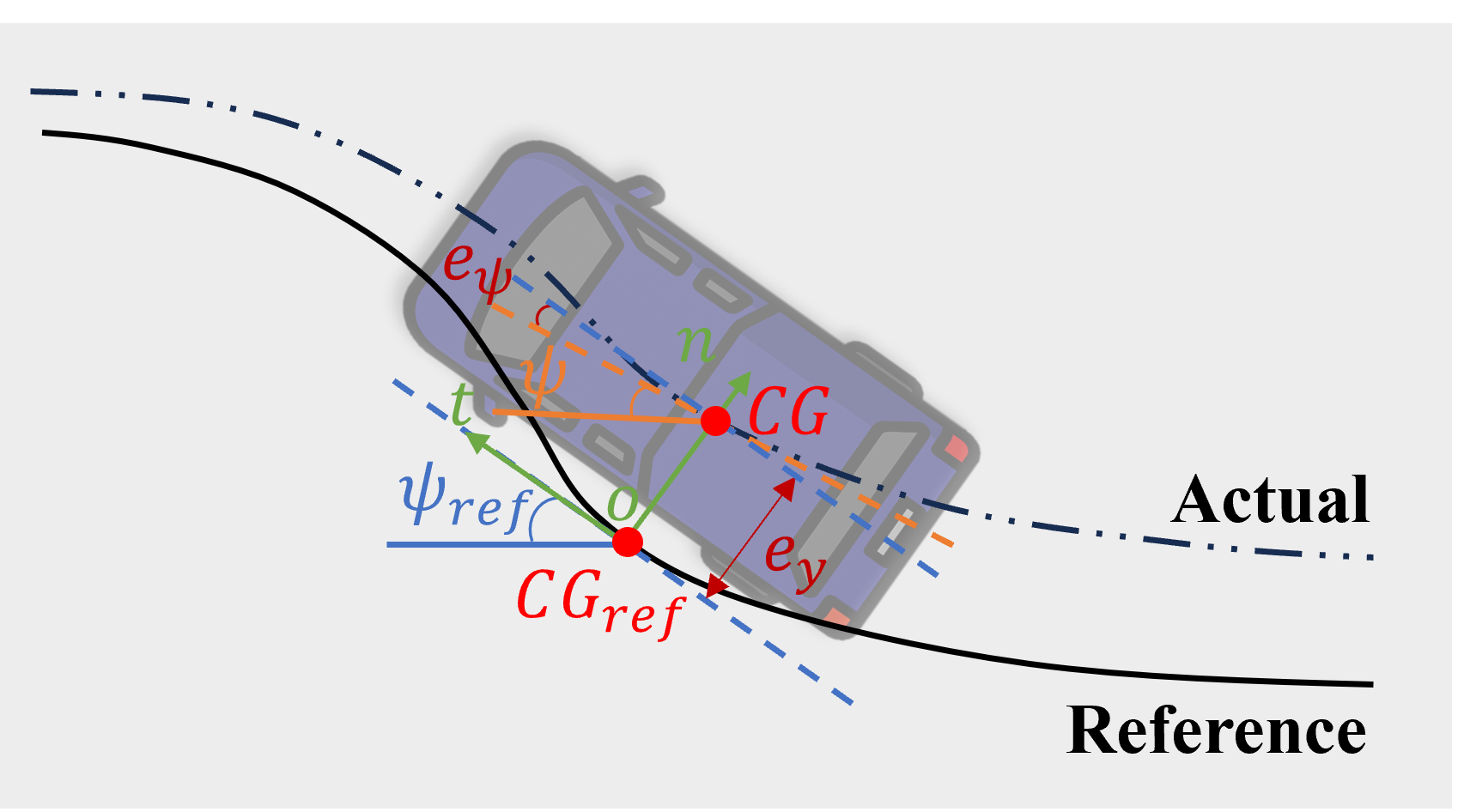}
	\caption{Path tracking kinematics model}
	\label{FIG:3}
\end{figure}

The schematic of vehicle path tracking is shown in \figref{FIG:3}, where the two red dots denote the actual position of $CG$ and the reference position $CG_{ref}$, respectively. The orange dashed line and the blue dashed line represent the tangents along the actual path and the reference path, respectively. A Frenet coordinate frame $o-tn$ is established on the reference path. The heading angle error $e_{\psi}$ is the difference between the vehicle actual heading angle $\psi$ and the tangent angle of the reference path $\psi_{ref}$ at $CG_{ref}$.
\par
Assuming that $e_{\psi}$ and $\dot\beta$ are both small, the path tracking kinematic model can be expressed as:
\begin{equation}\label{(15)}
\begin{cases} \dot{e}_{y} = v_{x}(\sin e_{\psi} + \beta \cos e_{\psi}) = v_{x}e_{\psi} + v_{x}\beta \\ \dot{e}_{\psi} = \dot{\psi}_{ref} - \dot{\psi} = \gamma - \kappa v_{x} \end{cases}.
\end{equation}
where $\kappa$ denotes the curvature of the reference path at the desired location of $CG$.
\subsection{UGV path tracking system}
By combining the vehicle dynamic model with the path tracking kinematic model, the state vector is selected as $\bm{x}=[e_y, e_{\psi}, \beta, \gamma]^T$, the input vector as $\bm{u}=[\delta, M_z]^T$, and the disturbance vector as $\bm\omega=[\sin\theta_B, \cos\theta_s, -\kappa{v_x}]^T$. The state-space equations of the UGV path tracking system can then be established and discretized in the following form:
\begin{equation}\label{(16)}
\bm{x}_{k+1} = \bm{A}(\bm{\tau}_{ij})\bm{x}_{k} + \bm{B}(\bm{\tau}_{ij})\bm{u}_{k} + \bm{E}\bm{\omega}_{k},
\end{equation}
where $\bm{A}(\bm{\tau}_{ij})=\bm{A}_0+\tau_{fl}\bm{A}_1+\tau_{fr}\bm{A}_2+\tau_{rl}\bm{A}_3+\tau_{rr}\bm{A}_4$, $\bm{B}(\bm{\tau}_{ij})=\bm{B}_0+\tau_{fl}\bm{B}_1+\tau_{fr}\bm{B}_2$, 
\par
\[\bm{A}_{0} = I + \Delta T \begin{bmatrix} 0 & v_{x} & v_{x} & 0 \\ 0 & 0 & 0 & 1 \\ 0 & 0 & -\frac{2(C_{f0} + C_{r0})}{mv_{x}} & \frac{2(C_{r0}l_{r} - C_{f0}l_{f})}{mv_{x}^{2}} - 1 \\ 0 & 0 & \frac{2(C_{r0}l_{r} - C_{f0}l_{f})}{I_{zz}} & -\frac{2(C_{f0}l_{f}^{2} + C_{r0}l_{r}^{2})}{I_{zz}v_{x}} \end{bmatrix},
\]
\par
\[
\bm{A}_1 = \bm{A}_2 = \Delta T \begin{bmatrix} 0 & 0 & 0 & 0 \\ 0 & 0 & 0 & 0 \\ 0 & 0 & -\frac{C_{fv}}{mv_x} & -\frac{C_{fv}l_f}{mv_x^2} \\ 0 & 0 & -\frac{C_{fv}l_f}{I_{zz}} & -\frac{C_{fv}l_f^2}{I_{zz}v_x} \end{bmatrix},\bm{A}_{3} = \bm{A}_{4} = \Delta T \begin{bmatrix} 0 & 0 & 0 & 0 \\ 0 & 0 & 0 & 0 \\ 0 & 0 & -\frac{C_{rv}}{mv_{x}} & \frac{C_{rv}l_{r}}{mv_{x}^{2}} \\ 0 & 0 & \frac{C_{rv}l_{r}}{I_{zz}} & -\frac{C_{rv}l_{r}^{2}}{I_{zz}v_{x}} \end{bmatrix},\]
\par
\[\bm{B}_{0} = \Delta T \begin{bmatrix} 0 & 0 \\ 0 & 0 \\ \frac{2 C_{f0}}{m v_{x}} & 0 \\ \frac{2 C_{f0} l_{f}}{I_{zz}} & \frac{1}{I_{zz}} \end{bmatrix},
\bm{B}_{1} = \bm{B}_{2} = \Delta T \begin{bmatrix} 0 & 0 \\ 0 & 0 \\ \frac{2 C_{fv}}{m v_{x}} & 0 \\ \frac{2 C_{fv} l_{f}}{I_{zz}} & \frac{1}{I_{zz}} \end{bmatrix},
\bm{E} = \Delta T \begin{bmatrix} 0 & 0 & 0 \\ 0 & 0 & 1 \\ \frac{g}{v_x} & 0 & 0 \\ 0 & \frac{d \mu m g (l_f - l_r)}{2 I_{zz} L} & 0 \end{bmatrix}.\]
where $\Delta{T}$ is the discrete time of system.

\section{Basic path tracking control strategy}\label{sec:3}

\begin{figure}
	\centering
		\includegraphics[width=0.9\textwidth]{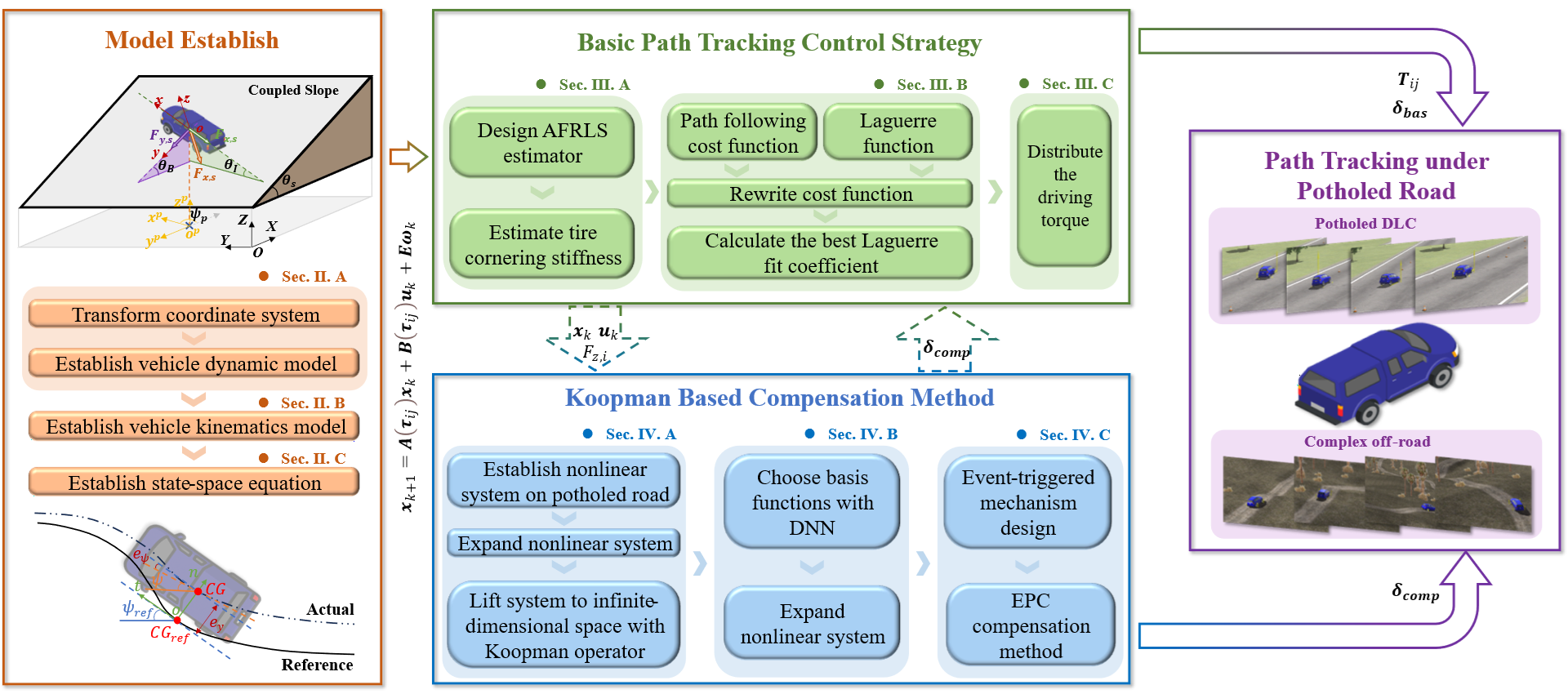}
	\caption{Path tracking control architecture on complex off-road scenario}
	\label{FIG:4}
\end{figure}

This paper proposes a path tracking control architecture for complex off-road scenarios, as shown in \figref{FIG:4}. Firstly, through coordinate frame transformations, a vehicle dynamic model and a vehicle kinematic model considering the coupled slope are constructed, thereby establishing a path tracking state-space equation for coupled slope scenarios that accounts for inconsistent four wheel cornering stiffness. Then, an AFRLS tire cornering stiffness estimation method is designed. On this basis, a baseline path tracking control strategy is developed by incorporating feasibility constraints, Laguerre functions, and the MPC path tracking cost function, yielding the baseline steering angle $\delta_\mathrm{bas}$ and the additional yaw moment $M_z$. Subsequently, based on Koopman operator theory, an offline DNN training method with an autoencoder structure is designed to obtain the optimal lifting function and the finite-dimensional approximation matrix, thereby producing the compensatory steering angle $\delta_\mathrm{comp}$. Finally, an EPC compensation mechanism is constructed via path deviation compensation activation criteria and credibility verification, which improves path tracking accuracy on potholed road while ensuring the vehicle’s handling reliability.
\subsection{Tire cornering stiffness estimate method}
In off-road scenarios, coupled slopes and potholed road cause continuous variations in tire vertical loads, leading to inconsistent and time-varying four wheel cornering stiffness $C_{ij}$. To accurately represent the tire cornering stiffness, this paper designs an AFRLS based tire cornering stiffness estimator to precisely estimate the time-varying stiffness correction coefficient $\bm{\tau}_k$ for four wheels,as shown in Algorithm \ref{alg:afr}.
\begin{algorithm}
\caption{AFRLS method}
\label{alg:afr}

\KwIn{initial time-varying stiffness correction coefficient $\bm{\tau}_0$, initial inverse covariance matrix $\bm{\Gamma}_0$, initial auxiliary variable $\eta_0$, Schur stable gain matrix $\bm{K}_e$, forgetting factor $\lambda$}
\KwOut{time-varying stiffness correction coefficient $\tau_{ij}$}

\While{$V \neq 0$}{
  \For{$k=0$ \KwTo $k_{\max}$}{
    Compute regression matrix $\bm{g}_k$ by \eqref{(18)}\;
    Update adaptive filter $\bm{p}_k$ by \eqref{(19)}\;
    Compute auxiliary variable $\bm{\eta}_k$ by \eqref{(22)}--\eqref{(23)}\;
    Update inverse covariance matrix by \eqref{(27)}\;
    Estimate time-varying stiffness correction coefficients $\hat{\boldsymbol{\tau}}_{i}$ by \eqref{(26)}\;
    Predict estimated state vector $\bm{\widehat{{x}}}_{k}$ by \eqref{(20)}\;
    }
}
\Return\
\end{algorithm}
\par
Firstly, a tire cornering stiffness estimator incorporating an adaptive filter is established. The state-space equations of the system are rewritten as follows.
\begin{equation}\label{(17)}
\bm{h}_{k} = \bm{g}_{k}\bm{\tau}_{k},
\end{equation}
\begin{equation}\label{(18)}
\boldsymbol{g}_{k} = [\boldsymbol{A}_{1}x_{k} + \boldsymbol{B}_{1}u_{k}, \boldsymbol{A}_{2}x_{k} + \boldsymbol{B}_{2}u_{k}, \boldsymbol{A}_{3}x_{k}, \boldsymbol{A}_{4}x_{k}],
\end{equation}
where $g_k$ is the regression matrix, the observation residual is $\bm{h}_k=\bm{x}_{k+1}-\bm{A}_0\bm{x}_k-\bm{B}_0\bm{u}_k-\bm{E}\bm{\omega}_k$, and the time-varying tire stiffness correction coefficient vector is $\bm{\tau}_k=[\tau_{fl,k}, \tau_{fr,k}, \tau_{rl,k}, \tau_{rr,k}]^T$.
When the vehicle travels straight, both the state vector $\bm{x}_k$ and the input vector $\bm{u}_k$ are approximately zero, resulting in $\bm{g}_k=0$. Thus, the conventional RLS method cannot guarantee the convergence of the tire cornering stiffness estimator. Therefore, this paper introduces an adaptive filter $\bm{p}_k$ to ensure persistent parameter updating and prevent estimation divergence:
\begin{equation}\label{(19)}
\bm{p}_{k+1} = \bm{g}_{k}-\bm{K}_{e}\bm{p}_{k},
\end{equation}
where, $\bm{K}_e$ is the Schur stable gain matrix, $\bm{p}_0=O$
\par
The tire cornering stiffness estimator is designed as follows:
\begin{equation}\label{(20)}
\bm{\widehat{{x}}}_{k+1} = \bm{A}_{0} \bm{x}_{k} + \bm{B}_{0} \bm{u}_{k} + \bm{E} \boldsymbol{\omega}_{k} + \bm{g}_{k} \widehat{\boldsymbol{\tau}}_{k} + \bm{K}_{e} \bm{\widetilde{{x}}}_{k} + \bm{p}_{k} (\widehat{\boldsymbol{\tau}}_{k} - \widehat{\boldsymbol{\tau}}_{k-1}),
\end{equation}
where, $\bm{\widehat{{x}}}_{k}$ is the estimated state vector at time step $k$, $\widehat{\boldsymbol{\tau}}_{k}$ is the estimate of the time-varying tire stiffness correction coefficient, and $\bm{\widetilde{{x}}}_{k}=\bm{x}_k-\bm{\widehat{{x}}}_{k}$ is the state estimation error.
\par
Then, a recursive least squares cost function with a forgetting factor is formulated. By combining \eqref{(19)} and \eqref{(20)}, the state estimation error $\bm{\widetilde{{x}}}_{k}=\bm{x}_k-\bm{\widehat{{x}}}_{k}$ can be obtained as:
\begin{equation}\label{(21)}
\bm{\tilde{x}}_{k+1} = \bm{g}_{k} \bm{\tilde{\tau}}_{k} - \bm{K}_{e} \bm{\tilde{x}}_{k} - \bm{p}_{k} (\bm{\hat{\tau}}_{k} - \bm{\hat{\tau}}_{k-1}).
\end{equation}
\par
To decouple the time-varying correction coefficient error $\bm{\tilde{\tau}}_{k}$ so as to ensure estimation feasibility, an auxiliary variable $\bm{\eta}_{k}$ is introduced to decompose $\bm{\tilde{x}}_{k}$ into the parameter-dependent term $\bm{\epsilon}_{k}$ that affects $\bm{\tilde{\tau}}_{k}$ and a parameter-independent error term $\bm{\eta}_{k}$:
\begin{equation}\label{(22)}
\bm{\eta}_{k} = \bm{\tilde{x}}_{k} - \bm{\epsilon}_{k},
\end{equation}
\begin{equation}\label{(23)}
\bm{\epsilon}_{k}=\bm{p}_k\bm{\tilde{\tau}}_{k}.
\end{equation}
\par
By treating $\bm{\eta}_{k}$ as the output, and by minimizing the mismatch between the estimated time-varying stiffness correction coefficients ${\hat{\boldsymbol{\tau}}}_{i}$ and the fitted constant error term $\bm{\tau}$, a forgetting recursive least squares cost function involving bilinear optimization can be formulated, and the nonlinear problem can be decoupled into:
\begin{equation}\label{(24)}
\begin{aligned}
\min_{\bm{\tau}, \bm{\hat{\tau}}_{i}} J_{k}(\boldsymbol{\tau})
&= \min_{\bm{\tau}, \bm{\hat{\tau}}_{i}} \sum_{i=0}^{k} \lambda^{k-i}
\left\lVert \boldsymbol{\epsilon}_{i} - \boldsymbol{p}_{i}\boldsymbol{\tau} \right\rVert^{2} \\
&= \min_{\bm{\tau}, \bm{\hat{\tau}}_{i}} \sum_{i=0}^{k} \lambda^{k-i}
\left\lVert \boldsymbol{p}_{i}\!\left(\boldsymbol{\tau}_{i}-\hat{\boldsymbol{\tau}}_{i}\right) - \boldsymbol{p}_{i}\boldsymbol{\tau} \right\rVert^{2},
\end{aligned}
\end{equation}
where, $\lambda \in (0, 1]$ is the forgetting factor.
\par
Finally, by minimizing the cost function $J_k(\tau)$, the tire cornering stiffness estimation law can be obtained as:
\begin{equation}\label{(25)}
\left( \sum_{i=0}^{k} \lambda^{k-i} \bm{p}_{i}^{T} \bm{p}_{i} \right) \hat{\bm{\tau}}_{k} = \sum_{i=0}^{k} \lambda^{k-i} \bm{p}_{i}^{T} \bm\epsilon_{i}.
\end{equation}
\par
By defining the inverse covariance matrix $\bm{\Gamma}_0$, the estimate of the time-varying tire stiffness correction coefficients $\hat{\tau}_{k}$ can be expressed in the following form:
\begin{equation}\label{(26)}
\widehat{\bm{\tau}}_{k} = \bm{\Gamma}_{k}^{-1} \left( \sum_{i=0}^{k} \lambda^{k-i} {p}_{i}^{T} \bm{\epsilon}_{i} \right),
\end{equation}
\begin{equation}\label{(27)}
\mathbf{\bm{\Gamma}}_{k} = \left( \sum_{i=0}^{k} \lambda^{k-i} \bm{p}_{i}^{T} \bm{p}_{i} \right) = \lambda \mathbf{\bm{\Gamma}}_{k-1} + \bm{p}_{k}^{T} \bm{p}_{k}
\end{equation}
\par
By combining \eqref{(26)}, \eqref{(27)} and performing algebraic manipulations, the following AFRLS-based tire cornering stiffness estimation method can be derived:
\begin{equation}\label{(28)}
\left\{
\begin{aligned}
\hat{\bm{\tau}}_{k+1} &= \hat{\bm{\tau}}_{k} + \bm{\Gamma}_{k+1}^{-1}\,\bm{p}_k^{T}\!\left(\bm{\tilde{x}}_k - \bm{\eta}_k\right)\\
\bm{\Gamma}_{k} &= \bm{\Gamma}_{k-1} + \bm{p}_k^{ T}\bm{p}_k
\end{aligned}.
\right.
\end{equation}

\subsection{Path tracking control strategy}
To ensure driving stability, riding comfort, and tracking accuracy, a multi-objective optimization problem for UGV path tracking is formulated as:
\begin{equation}\label{(29)}
\min_{\Delta u_k} J = \sum_{i=1}^{N_p} \| x_{k+i} - x_{\mathrm{ref},k+i} \|_Q^2 + \| \Delta u_{k+i} \|_R^2,
\end{equation}
\begin{equation}\label{(30)}
\begin{aligned}
\text{s.t.}\;\; &
\vcenter{\hbox{$
\left\{
\begin{array}{c}
\min_{\Delta u_k} J \\
\left|\Delta u_{k+i}\right| \le \Delta u_{\lim} \\[2pt]
\left|u_{k+i}\right| \le u_{\lim}
\end{array}
\right.
$}}
\end{aligned},
\end{equation}
where $x_{\mathrm{ref}}$ is the reference state vector, $\Delta u_{k+i}$ is the input increment, $\Delta u_{\mathrm{lim}}$ is the maximum input increment, $u_{\mathrm{lim}}$ is the maximum input, $Q$ is the output weighting matrix, and $R$ is the input weighting matrix.
\par
In conventional MPC, quadratic programming is typically used to solve the cost function and obtain the optimal input sequence. However, quadratic programming requires substantial computational resources and cannot satisfy the real-time requirements during vehicle operation. Therefore, we employ Laguerre function to fit the optimal input, replacing the quadratic-programming step in traditional MPC and reducing the computational burden of the control strategy.
\par
\begin{equation}\label{(31)}
\left\{
\begin{gathered}
\xi_{1}(z)=\dfrac{\sqrt{1-\alpha^{2}}}{1-\alpha z^{-1}}\\[6pt]
\xi_{2}(z)=\dfrac{\sqrt{1-\alpha^{2}}}{1-\alpha z^{-1}}
           \dfrac{z^{-1}-\alpha}{1-\alpha z^{-1}}\\[6pt]
\vdots \\[6pt]
\xi_{N}(z)=\dfrac{\sqrt{1-\alpha^{2}}}{1-\alpha z^{-1}}
           \left(\dfrac{z^{-1}-\alpha}{1-\alpha z^{-1}}\right)^{N-1}
\end{gathered},
\right.
\end{equation}
where $\xi_{i}(z)$ denotes the discrete Laguerre function, $0 \leq \alpha \leq 1$ is the pole of the discrete Laguerre function, and $N$ is the number of the discrete Laguerre function terms.
\par
To simplify the analysis of the discrete UGV path tracking system, the $z$-transform is applied to $\xi_{i}(z)$, so that the discrete Laguerre function can be expressed in the following form:
\begin{equation}\label{(32)}
\wp_{k} = [l_{1,k}, l_{2,k}, \dots, l_{N,k}]^{T},
\end{equation}
where $l_{i,k}$ is the $z$-transform of $\xi_{i}(z)$ and forms the polynomial of the discrete Laguerre functions, $\wp_{k}$ is the path tracking Laguerre polynomial designed in this paper.
\par
By combining \eqref{(31)} and \eqref{(32)}, the path tracking Laguerre polynomial can be expressed recursively as:
\begin{equation}\label{(33)}
\wp_{k+1} = \Upsilon \wp_{k},
\end{equation}
\begin{equation}\label{(34)}
\Upsilon = \begin{bmatrix} \alpha & 0 & 0 & \dots & 0 \\ \beta & \alpha & 0 & \dots & 0 \\ -\alpha \beta & \beta & \alpha & \dots & 0 \\ \vdots & \vdots & \vdots & \ddots & \vdots \\ (-\alpha)^{N-2} \beta & (-\alpha)^{N-3} \beta & (-\alpha)^{N-4} \beta & \dots & \alpha \end{bmatrix},
\end{equation}
\begin{equation}\label{(35)}
\wp_{0} = \sqrt{\beta}[1 \quad -\alpha \quad \alpha^{2} \quad -\alpha^{3} \cdots (-1)^{N-1} \alpha^{N-1}],
\end{equation}
where $\Upsilon$ is an $N \times N$ recurrence matrix, $\beta=1-\alpha^2$.
\par
On this basis, the control input $u_{k+i}$ can be represented using the path tracking Laguerre polynomial as:
\begin{equation}\label{(36)}
u_{k+i} = \sum_{j=1}^{N} c_{j,k} l_{j,i} = \wp_{k+i}^{T} \eta,
\end{equation}
\begin{equation}\label{(37)}
\eta = \begin{bmatrix} c_{1,k}(\Delta \delta_\mathrm{bas}), c_{2,k}(\Delta \delta_\mathrm{bas}), ..., c_{N,k}(\Delta \delta_\mathrm{bas})\\ c_{1,k}(M_{z}), c_{2,k}(M_{z}), ..., c_{N,k}(M_{z}) \end{bmatrix}^{T},
\end{equation}
where $c_{j,k}(\Delta \delta_\mathrm{bas})$ and $c_{j,k}(M_{z})$ are the Laguerre fitting coefficients for the baseline steering angle increment and the additional yaw moment, respectively.
\par
Then, the discrete state-space equations and the cost function are reformulated using the baseline steering angle and additional yaw moment represented by the path tracking Laguerre polynomial. By substituting the input $u_k$ into the discrete state-space equations and performing the following iterative derivation in conjunction with \eqref{(36)}, the system state and output expressed in terms of the path tracking Laguerre polynomial can be obtained as:
\begin{equation}\label{(38)}
\left\{
\begin{aligned}
x_{k+1} &= \bar{A}x_k + \bar{B}\,\wp_k^{T}\eta + \bar{E}\,\omega_k,\\
x_{k+2} &= \bar{A}x_{k+1} + \bar{B}\,\wp_{k+1}^{T}\eta + \bar{E}\,\omega_{k+1} \\
        &= \bar{A}^{2}x_k + \bar{A}\bar{B}\,\wp_k^{T}\eta + \bar{A}\bar{E}\,\omega_k
           + \bar{B}\,\wp_{k+1}^{T}\eta + \bar{E}\,\omega_{k+1}\\
&\ \ \vdots \\
x_{k+i} &= \bar{A}^{i}x_k +  \sum_{j=0}^{i-1}\bar{A}^{\,i-1-j}\bar{B}\,\wp_{k+j}^T\eta
\end{aligned}.
\right.
\end{equation}
\par
Similarly, the cost function of the vehicle path tracking optimization problem can be reformulated using the path tracking Laguerre polynomial as:
\begin{equation}\label{(39)}
\begin{aligned}
J^{*}
= \sum_{i=1}^{N_{p}}
\Big(
\vartheta^{T} Q \Omega \eta
+ \eta^{T}\Omega^{T} Q \vartheta
+ \eta^{T}\Omega^{T} Q \Omega \eta + \eta^{T}\!\left(\wp^{T}_{k+i}-\wp^{T}_{k+i-1}\right)
R\left(\wp^{T}_{k+i}-\wp^{T}_{k+i-1}\right)^{T}\Big)
\end{aligned}.
\end{equation}
\begin{equation}\label{(40)}
\Omega = \sum_{j=0}^{i-1} \bar{A}^{i-j-1} \bar{B} \wp_{k+j}^{T},
\end{equation}
\begin{equation}\label{(41)}
\vartheta = \overline{A}^{i} x_{k}.
\end{equation}
\par
Finally, the optimal input sequence for path tracking is obtained by differentiating with respect to the Laguerre fitting coefficient $\eta$. The objective of the path tracking control in this paper is to determine an appropriate input sequence that minimizes the established cost function. Through the above transformations, the cost function $J^*$ has been expressed as a function of the Laguerre fitting coefficient $\eta$. Therefore, by setting the derivative of the cost function with respect to $\eta$ to zero, the corresponding optimal Laguerre fitting coefficient can be obtained as:
\begin{equation}\label{(42)}
\begin{aligned}
\frac{\partial J^{*}}{\partial \eta}
= 2\sum_{i=1}^{N_{p}} \Omega^{T}Q^{T}\vartheta
   + \Omega^{T}Q^{T}\Omega\,\eta + \left(\wp_{k+i}-\wp_{k+i-1}\right)
R\left(\wp_{k+i}-\wp_{k+i-1}\right)^{T}\eta ,
\end{aligned}
\end{equation}
\begin{equation}\label{(43)}
\eta_{\mathrm{opt}} = -\kappa^{-1}\times \varpi,
\end{equation}
where $\eta_{\mathrm{opt}}$ is the optimal Laguerre fitting coefficient, $\kappa$ and $\varpi$ are $N \times N$ and $N \times 1$ matrices, respectively.
\begin{equation}\label{(44)}
\begin{aligned}
\kappa
= \sum_{i=1}^{N_p}\Big(\left(\wp_{k+i}-\wp_{k+i-1}\right)
R\left(\wp_{k+i}-\wp_{k+i-1}\right)^{T} + \Omega^{T}Q\Omega\Big)
\end{aligned},
\end{equation}
\begin{equation}\label{(45)}
\varpi
= \sum_{i=1}^{N_p}\Omega^{T}Q^{T}\vartheta .
\end{equation}
\par
Based on the above derivations and in conjunction with \eqref{(36)}, the input sequence fitted by the Laguerre function can be obtained. In each control cycle, the first element of the sequence is selected, yielding the optimal baseline steering angle increment $\Delta\delta_{\mathrm{bas}}$ and the additional yaw moment $M_z$ for path tracking.

\subsection{Distribution of driving torque}
In this section, quadratic programming is employed to solve for the additional driving torque of each wheel. Since reducing tire utilization can improve the stability of vehicle path tracking while ensuring the available adhesion capability of each tire, the additional driving torque is optimally allocated by minimizing the sum of the squared adhesion utilization rates of the four tires. The objective function is formulated as:
\begin{equation}\label{(46)}
\min_{T_{i,j}}\;  J_T
= \sum \frac{T_{ij}^{2}}{\left(\mu F_{z,ij} r\right)},
\end{equation}
where $r$ is the wheel rolling radius.
\par
To ensure that the allocation of additional driving torques satisfies the required additional yaw moment $M_z$ while keeping the torque of each tire within its friction limit, the following equality and inequality constraints are defined as:
\begin{equation}\label{(47)}
\begin{aligned}
\text{s.t.}\;\; &
\left\{
\begin{aligned}
\frac{2M_z r}{d}
&= (T_{fr}-T_{fl})\cos\delta + (T_{rr}-T_{rl})\\
|T_{ij}|
&\le \min\!\left(\mu F_{z,ij},\; T_{\max}\right)
\end{aligned}
\right.
\end{aligned},
\end{equation}
where $T_{\mathrm{max}}$ is the maximum motor output torque.

\section{Path deviation compensation based on DK}\label{sec:4}
When the UGV travels on a smooth coupled slope, the vehicle can accurately track the desired path with LMPC. However, off-road scenarios involve complex ground conditions, such as potholes, that degrade path tracking accuracy. Therefore, this section designs a DNN Koopman operator based path deviation compensation method to generate a compensatory steering angle for accurate path tracking on potholed road.
\subsection{Data-Driven theory based on Koopman operator}
When the vehicle passes over potholes of different scales, the vertical loads on different wheels are different, which directly affects the degree of path deviation. Therefore, an auxiliary input $U_k$ is constructed, consisting of the four-wheel vertical loads $F_{z,ij}$, the compensatory steering angle $\delta_\mathrm{bas}$, and the additional yaw moment $M_z$. Meanwhile, to predict the degree of vehicle path deviation, the compensatory steering angle $\delta_\mathrm{comp}$ is incorporated into the original discrete state $x_k$ to form an auxiliary discrete state $X_k$. The discrete nonlinear path tracking dynamical system when the vehicle traverses potholed road is given by:
\begin{equation}\label{(48)}
X_{k+1}=f(X_k, U_k)
\end{equation}
where $f: \mathbb{R}^{N_X} \times \mathbb{R}^{N_U} \rightarrow \mathbb{R}^{N_X}$ is the nonlinear mapping function of system.
\par
To predict the compensatory steering angle of the vehicle, this paper employs Koopman operator theory extended to controlled systems to re-parameterize the nonlinear vehicle dynamical system.
\par
First, a state space $\mathcal{X}$ is defined as the product of the auxiliary discrete-state space $X_k$ and the space of all auxiliary inputs $U_k$, i.e., $\mathbb{R}^{N_X} \times \ell(U)$. Here, $\mathbb{R}^{N_X} \times \ell(U)$ denotes the space composed of all control sequences $U = \{U(i)\}_{i=0}^{\infty}$, with $U(i)\in \mathcal{U}$. Then, the extended system of \eqref{(48)} in this state space can be expressed as:
\begin{equation}\label{(49)}
\mathcal{X}_{k+1} = F(\mathcal{X}_{k}) = \begin{bmatrix} f({X}_{k}, {U}_{k}(0)) \\ \mathcal{L} {U}_{k} \end{bmatrix}
\end{equation}
where $\mathcal{X}_k=[X_k, U_k]^T$ is the extended state vector, $F: \mathbb{R}^{N_X} \times \ell(U)\rightarrow \mathbb{R}^{N_X}\times \ell(U)$, $\mathcal{L}$ is the left shift operator for updating the control sequence, i.e., $\mathcal{L}U_k=U_{k+1}$. $U_k(0)\in \mathbb{R}^{N_U}$ is the first element of the front wheel steering angle sequence $U$ at time step $k$.
\par
Next, for the extended system \eqref{(49)}, the Koopman operator $\mathcal{K} : \mathcal{H} \rightarrow \mathcal{H}$ is defined in an infinite-dimensional Hilbert space $\mathcal{H}$ as an infinite-dimensional linear operator acting on the observable function $\phi$:
\begin{equation}\label{(50)}
\mathcal{K} \phi(\mathcal{X}_{k}) = \phi(\mathcal{X}_{k+1}) = \phi(F(\mathcal{X}_{k})),
\end{equation}
where $\phi:\mathbb{R}^{N_X}\times \ell(U)\rightarrow \mathbb{R}^{N_X}\times \ell(U)$ is the observable function.
\par
Finally, under the action of the observable function $\phi$, the discrete nonlinear path tracking dynamical system \eqref{(48)} is lifted to the infinite-dimensional space $\mathcal{H}$, where its linear evolution is governed via the Koopman operator $\mathcal{K}$.
\subsection{Deviation compensation based on DK method}
For applying the infinite-dimensional Koopman operator $\mathcal{K}$ to the practical computation of the vehicle steering angle, this paper adopts the DK method to obtain a finite-dimensional approximation matrix of the Koopman operator.
\par
First, the observable function $\phi$ is assumed to have the form:
\begin{equation}\label{(51)}
\phi({\mathcal{X}}_{k}) = \begin{bmatrix} {\psi}(X_{k}) \\ U_{k} \end{bmatrix},
\end{equation}
where $\phi \in \mathbb{R}^{N_\phi}$, and $\psi(X_{k})=[\psi_1(X_k) \cdots \psi_{N_Z}(X_k)]^T$ is the vector of lifting functions, with $\psi_{i} : \mathbb{R}^{N_{X}} \rightarrow \mathbb{R}, i \in \{1, \cdots, N_{z}\}$. $N_z=N_{\phi}-N_U$ is the lifted-state dimension.
\par
Accordingly, \eqref{(50)} can be simplified as:
\begin{equation}\label{(52)}
\begin{bmatrix} \psi(X_{k+1}) \\ {U}_{k+1} \end{bmatrix} = K \begin{bmatrix} \psi(X_{k}) \\ {U}_{k} \end{bmatrix},
\end{equation}
where $K \in \mathbb{R}^{N_\phi \times N_\phi}$ is the finite-dimensional approximation matrix of the Koopman operator $\mathcal{K}$.
\par
To facilitate the computation of the finite-dimensional approximation matrix $K$, $[\bar{A} \ \bar{B}]$ is defined as the first $N_z$ rows of $K$, where $\bar{A}\in \mathbb{R}^{N_z \times N_z}$ and $\bar{B}\in \mathbb{R}^{N_z \times N_U}$.
\par
\begin{figure}
	\centering
		\includegraphics[width=0.6\textwidth]{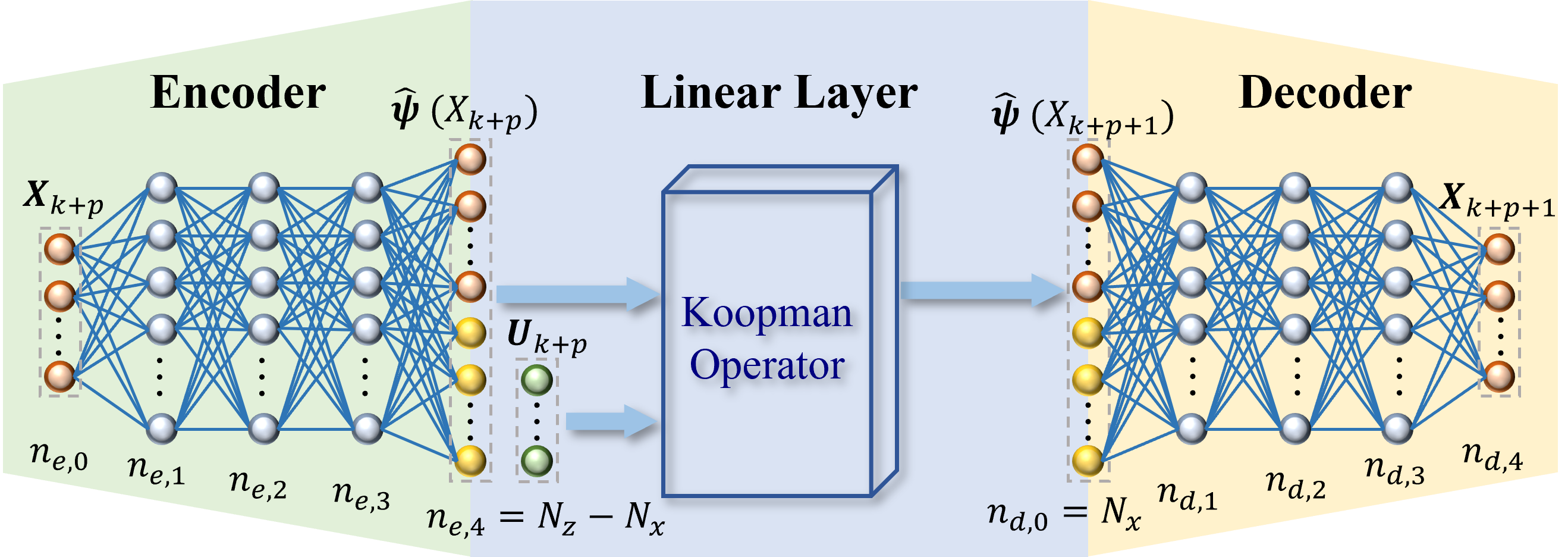}
	\caption{DK method learning framework}
	\label{FIG:5}
\end{figure}

Then, the lifting-function vector $\psi$ is constructed. In the traditional EDMD method, $\psi$ is constructed by manually selecting basis functions such as Gaussian functions and radial basis functions, which makes it difficult for the Koopman operator to achieve a good approximation. To improve the fitting accuracy of the vehicle compensatory steering angle, we construct a DNN to obtain the optimal $\psi$, as shown in \figref{FIG:5}. To facilitate the reconstruction of the auxiliary discrete state $X_{k+p}$ during lifting and reduction, $\psi$ is designed to include $X_{k+p}$ and other lifted observables. The encoder (En) and decoder (De) are both composed of four fully connected layers, and the numbers of neurons in the $i$-th layer of the encoder and decoder are $n_{e,i}$ and $n_{d,i}$, respectively. Rectified linear unit (ReLU) activation functions are used in the hidden layers to balance the accuracy of nonlinear mapping and computational efficiency, and no activation function is used in the output layer to preserve the linear property of the Koopman operator and the state reconstruction accuracy.
To ensure the accuracy of the learned vector of lifting functions for predicting the degree of vehicle path deviation, a loss function incorporating multi-dimensional errors is designed.
\par
On the one hand, to guarantee the prediction accuracy of the Koopman operator for the auxiliary vehicle state, a linear-error loss, which reflects the linear property of the Koopman operator in the high-dimensional space, and a multi-step state-prediction error loss, which reflects the state prediction accuracy, are formulated as:
\begin{equation}\label{(53)}
L_{lin} = \frac{1}{p} \sum_{i=1}^{p} ||\widehat{\psi}(X_{k+i}) - \psi(X_{k+i})||_{2}^{2},
\end{equation}
\begin{equation}\label{(54)}
L_{pre} = \frac{1}{p} \sum_{i=1}^{p} ||\widehat{X}_{k+i} - X_{k+i}||_{2}^{2},
\end{equation}
\begin{equation}\label{(55)}
\left\{
\begin{aligned}
\widehat{\boldsymbol{\psi}}(x_{k+p}) &= \overline{A} \boldsymbol{\psi}(X_{k+p-1}) + \overline{B} {U}_{k+p-1}\\
&= \overline{A}^{2} \boldsymbol{\psi}(X_{k+p-2}) + \overline{A} \overline{B} {U}_{k+p-2} + \overline{B} {U}_{k+p-1}\\
&\ \ \vdots \\
&= \overline{A}^{p} \boldsymbol{\psi}(X_{k}) + \sum_{i=1}^{p} \overline{A}^{i-1} \overline{B} {U}_{k+p-i},
\end{aligned}
\right.
\end{equation}
where $p$ is the prediction horizon, and $\widehat{X}_{k+i}=\mathrm{De}(\widehat{\psi}(X_{k+i}))$.
\par
On the other hand, to ensure accuracy in state reconstruction, a reconstruction-error loss $L_{rec}$ is defined as:
\begin{equation}\label{(56)}
L_{rec} = \| \mathrm{De}[\psi(X_{k})] - X_{k} \|_{2}^{2}.
\end{equation}
Accordingly, the total loss function $L_{tot}$ which composed of the above loss terms is given by:
\begin{equation}\label{(57)}
L_{tot} = \omega_1 L_{lin} + \omega_2 L_{pre} + \omega_3 L_{rec},
\end{equation}
where $\omega_1 - \omega_3$ are the weighting coefficients of the respective loss terms.
\par
By training the above DNN, the optimal $\psi$ can be obtained.
\par
Finally, based on $N$ sets of pre-collected vehicle states and control inputs, the nominal analytical solution of $[\bar{A} \ \bar{B}]$ is obtained by minimizing the error between the actual states and the predicted states:
\begin{equation}\label{(58)}
\min_{A,B} \sum_{j=1}^{N} \left\| \psi(X_{k+1}^{j}) - \bar{A} \psi(X_{k}^{j}) - \bar{B} U_{k}^{j} \right\|_{2}^{2}.
\end{equation}
Based on the trained $\bar{A}$ and $\bar{B}$, the finite-dimensional approximation matrix of the Koopman operator can be constructed as $K=[\bar{A} \ \bar{B}]$. According to \eqref{(52)}, the lifted function vector $\psi(X_k)$ can be predicted which include the compensatory steering angle $\delta_\mathrm{comp}$.
\subsection{Event-Triggered parallel cooperative compensation path tracking}
The vehicle can achieve higher-accuracy tracking of the desired path after compensating for vehicle path deviation with the DK method. However, on the one hand, the compensatory steering angle $\delta_\mathrm{comp}$ obtained by the DK method lacks clear interpretability during its generation. Directly superimposing it on the baseline steering angle produced by LMPC is not reliable. It may violate the constraints of the vehicle steering system, thereby deteriorating path tracking performance and even causing instability. On the other hand, when road excitation is mild, the DK method still consumes computational resources while providing no significant improvement. Therefore, this paper designs an EPC compensation mechanism to improve path tracking accuracy while ensuring vehicle handling reliability, as shown in Algorithm 2:
\begin{algorithm}
\caption{EPC compensation mechanism}
\label{alg:epc}

\KwIn{state vector $x_k$, input vector $u_k$, vertical load $F_{z,i}$, Koopman approximation matrix $K$}
\KwOut{total steering angle $\delta_{\mathrm{tot}}$}

\While{$V \neq 0$}{
  \For{$k=0$ \KwTo $k_{\max}$}{
    Estimate tire cornering stiffness correction coefficient $\tau_{i,j}$ by \eqref{(17)}--\eqref{(28)}\;
    Compute baseline steering angle $\delta_{\mathrm{bas}}$ by \eqref{(31)}--\eqref{(45)}\;
    Compute load transfer rate $\mathrm{LTR}$ by \eqref{(59)}\;

    \If{$\mathrm{LTR} \ge |\mathrm{LTR}|_{\mathrm{thr}}$}{
      Compute compensatory steering angle $\delta_{\mathrm{comp}}$ by \eqref{(52)}\;
      Compute credibility coefficient by \eqref{(61)}\;
    }
    \If{$\mathrm{LTR} < |\mathrm{LTR}|_{\mathrm{thr}}$}{
      $\delta_{\mathrm{comp}} \leftarrow 0$\;
    }

    Compute total steering angle $\delta_{\mathrm{tot}}$ by \eqref{(60)}\;
  }
}
\Return\
\end{algorithm}
\par
First, a path deviation compensation activation criterion is designed. When the UGV is operating on a coupled slope under straight driving and cornering conditions, the load transfer rate (LTR) is typically below 0.3 \cite{35Zhang2025}. When traversing potholed road, the LTR of the UGV is generally greater than 0.6, and it can directly reflect the pothole scale and the degree of tracking deviation \cite{18Liu20242}:
\begin{equation}\label{(59)}
\mathrm{LTR} = \frac{F_{z,fr} + F_{z,rr} - F_{z,fl} - F_{z,rl}}{F_{z,fr} + F_{z,rr} + F_{z,fl} + F_{z,rl}}.
\end{equation}
\par
When the LTR remains low, the UGV exhibits only minor deviation from the reference path, and LMPC is sufficient to maintain accurate path tracking. When the LTR is large, a significant deviation occurs between the vehicle and the desired path. The UGV achieves path tracking by combining the baseline steering angle $\delta_\mathrm{bas}$ generated by the LMPC with the compensatory steering angle $\delta_\mathrm{comp}$ provided by the DNN Koopman method.
\begin{equation}\label{(60)}
\delta_{tot} = \begin{cases} \delta_\mathrm{bas}, & |\mathrm{LTR}| \leq \mathrm{LTR}_\mathrm{thr} \\ \delta_\mathrm{bas} + \lambda \delta_\mathrm{comp}, & |\mathrm{LTR}| > \mathrm{LTR}_\mathrm{thr} \end{cases},
\end{equation}
where $\delta_{tot}$ is the total steering angle, $\mathrm{LTR}_\mathrm{thr}$ is the LTR triggering threshold, and $\lambda$ is the credibility coefficient used to ensure the reliability of the total steering command.
\par
Then, the EPC compensation mechanism is designed. When the vehicle traverses large pothole surfaces and satisfies the activation criterion in \eqref{(60)}, the DNN Koopman method generates a compensatory steering angle based on the vehicle states. Subsequently, the generated compensatory steering angle $\delta_\mathrm{comp}$ is fed back to the LMPC module for credibility verification, to determine the credibility coefficient $\lambda$, as shown in \figref{FIG:6}:
\begin{figure}
	\centering
		\includegraphics[width=0.5\textwidth]{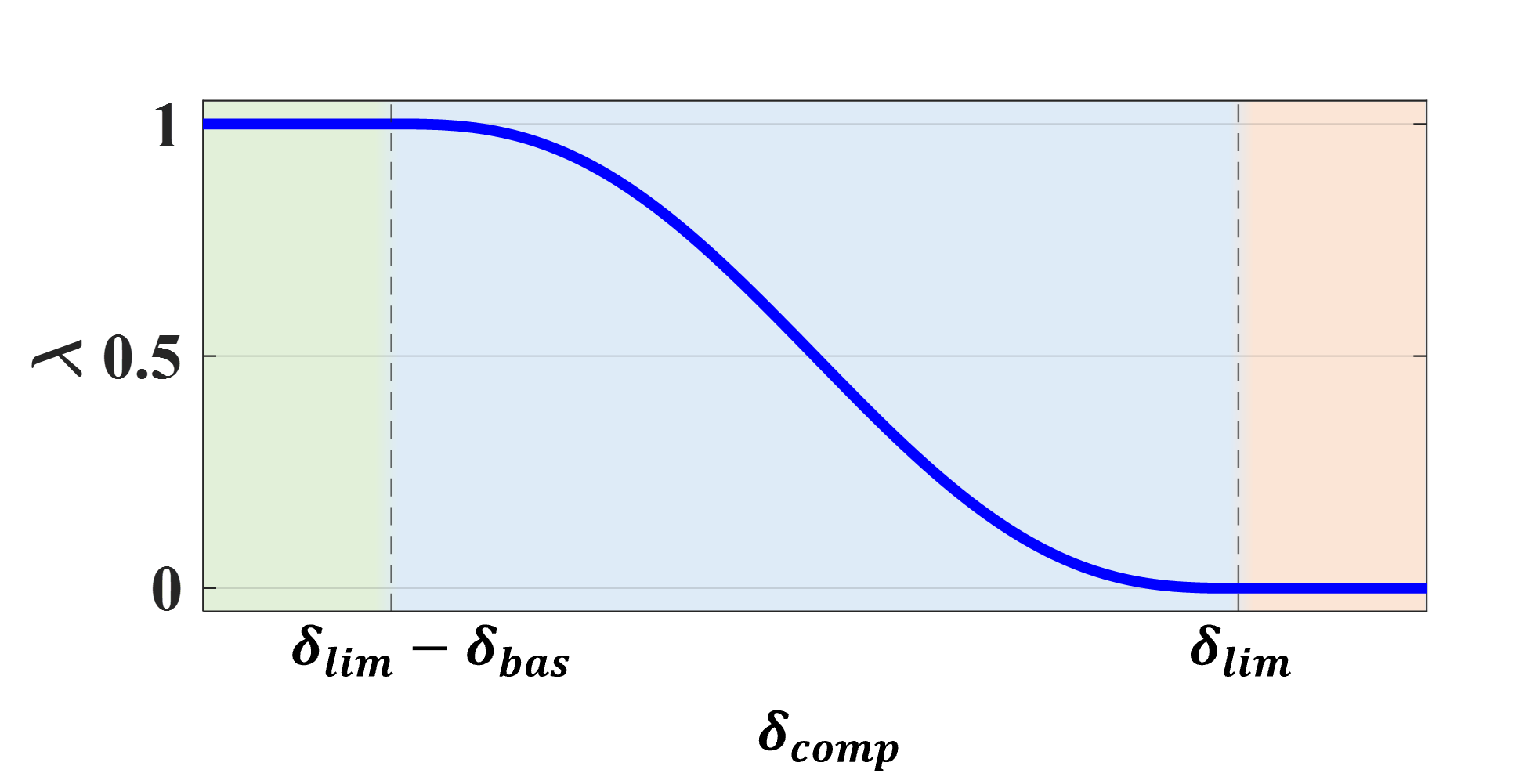}
	\caption{Credibility coefficient}
	\label{FIG:6}
\end{figure}
\begin{equation}\label{(61)}
\lambda = \begin{cases} 1, & \delta_{\mathrm{comp}} \leq \delta_0 \\ \frac{1}{1 + e^{5(\frac{\delta_{\mathrm{comp}} - \delta_0}{\delta_1 - \delta_0} - 0.5)}}, & \delta_0 \leq \delta_{\mathrm{comp}} \leq \delta_1 \\ 0, & \delta_1 \leq \delta_{\mathrm{comp}} \end{cases},
\end{equation}
\begin{equation}\label{(62)}
\begin{cases} \delta_{0} = \delta_{\mathrm{lim}} - \delta_{\mathrm{bas}} \\ \delta_{1} = \delta_{\mathrm{lim}} \end{cases},
\end{equation}
where $\delta_{\mathrm{lim}}$ is the maximum front wheel steering angle of the vehicle.
\par
When the sum of the compensatory steering angle $\delta_{\mathrm{comp}}$ and the baseline steering angle is still smaller than $\delta_{\mathrm{lim}}$, $\delta_{\mathrm{comp}}$ is regarded as fully reliable and is entirely used for path deviation compensation. When $\delta_{\mathrm{comp}} < \delta_{\mathrm{lim}}$ but the superposition with the baseline steering angle exceeds the physical constraint, $\delta_{\mathrm{comp}}$ is considered partially reliable, and only part of it is used to compensate the path deviation. When $\delta_{\mathrm{comp}} > \delta_{\mathrm{lim}}$, the generated compensatory steering angle is considered completely unreliable. In this case, UGV path tracking relies only on $\delta_{\mathrm{bas}}$ to maintain vehicle stability.
\par
With the constraint in \eqref{(61)}, the total steering angle is guaranteed to satisfy the physical constraints of the steering system, thereby improving path tracking accuracy while ensuring vehicle handling reliability.

\section{Simulation results and discussion}\label{sec:5}
\subsection{Experimental platform and cases}
\begin{figure}
	\centering
		\includegraphics[width=0.55\textwidth]{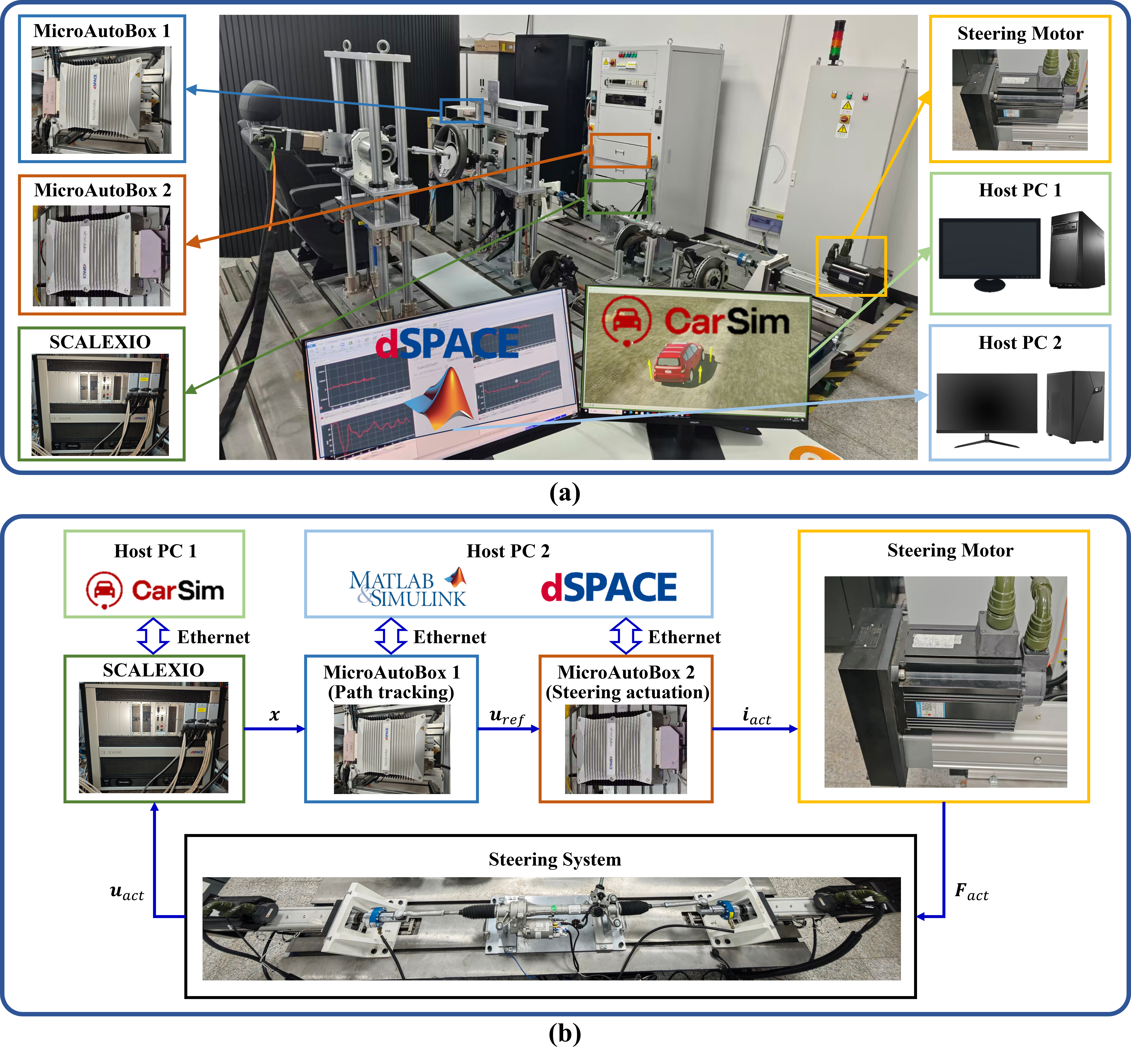}
	\caption{HiL platform of path tracking}
	\label{FIG:7}
\end{figure}

In this section, a HiL experimental platform is built, as shown in \figref{FIG:7}, and experiments are conducted for validation.
The HiL platform consists of Host PC 1, SCALEXIO, Host PC 2, MicroAutoBox 1, MicroAutoBox 2, steering motors, and steering actuation system. Signals among the components are transmitted and received via the CAN bus. Host PC 1 deploys the compiled CarSim model to SCALEXIO via Ethernet. Similarly, the compiled path tracking control strategy and the steering motor control strategy on Host PC 2 are deployed to MicroAutoBox 1 and MicroAutoBox 2 via Ethernet, respectively. MicroAutoBox 1 receives the vehicle states $x$ sent by SCALEXIO through the CAN bus and generates the desired input $u_\mathrm{ref}$
 using the path tracking control strategy. The steering motor control strategy in MicroAutoBox 2 generates the actual control current $i_\mathrm{act}$ to track $u_\mathrm{ref}$. The steering motor actuates according to $i_\mathrm{act}$, producing the actual input $u_\mathrm{act}$ through the complete physical steering system, which is then fed back as the input to the CarSim model running on SCALEXIO. The HiL platform is monitored and data are collected using the ControlDesk software running on the host PC.
\par
This paper selects the D-Class SUV V9 in CarSim as the experimental vehicle, and the related vehicle parameters are listed in TABLE~\ref{tab:1}. The parameters of the UGV path tracking control strategy are given in TABLE~\ref{tab:2}.
\begin{table}
\caption{Parameters of vehicle}
\label{tab:1}
\centering
\begin{tabular}{lcc}
\toprule
\textbf{Parameter} & \textbf{Notation} & \textbf{Value} \\
\midrule
Vehicle mass                    & $m$     & $1430\,\mathrm{kg}$ \\
Yaw moment of inertia           & $I_{zz}$ & $2059\,\mathrm{kg\cdot m^{2}}$ \\
Distance from CG to front axle  & $l_f$   & $1.050\,\mathrm{m}$ \\
Distance from CG to rear axle   & $l_r$   & $1.610\,\mathrm{m}$ \\
Wheel spacing                   & $d$     & $1.565\,\mathrm{m}$ \\
Height of CG                    & $h_g$   & $0.65\,\mathrm{m}$ \\
Tire rolling radius             & $r$     & $0.325\,\mathrm{m}$ \\
Nominal cornering stiffness     & $C_{i0}$ & $-850\,\mathrm{N/deg}$ \\
Cornering stiffness variation range & $C_{iv}$ & $-1000\,\mathrm{N/deg}$ \\
\bottomrule
\end{tabular}
\end{table}

\begin{table}
\caption{Parameters of control algorithm}
\label{tab:2}
\centering 
\begin{tabular}{lcc}
\toprule
\textbf{Parameter} & \textbf{Notation} & \textbf{Value} \\ 
\midrule
Control period & $T_s$ & $0.01\,\mathrm{s}$ \\
Prediction horizon & $N_p$ & $20$ \\
Control horizon & $N_c$ & $10$ \\

Output weight matrix & $Q$ & $\mathrm{diag}[1000,500,1,1]$ \\
Input weight matrix & $R$ & $\mathrm{diag}[10,1]$ \\
Load transfer rate threshold & $\mathrm{LTR}_{\mathrm{thr}}$ & $0.6$ \\
Number of encoder neurons & $n_{e,1\sim4}$ & $(11,64,128,64)$ \\
Number of decoder neurons & $n_{d,1\sim4}$ & $(5,64,128,64)$ \\
\bottomrule
\end{tabular}
\end{table}
To verify the effectiveness of the proposed UGV path tracking control strategy based on cooperative DK–LMPC (CDK–LMPC), two test cases are designed as follows:
\\
\textbf{Case 1: Double lane change on a coupled slope with potholes}
\par
\begin{figure}
	\centering
		\includegraphics[width=0.7\textwidth]{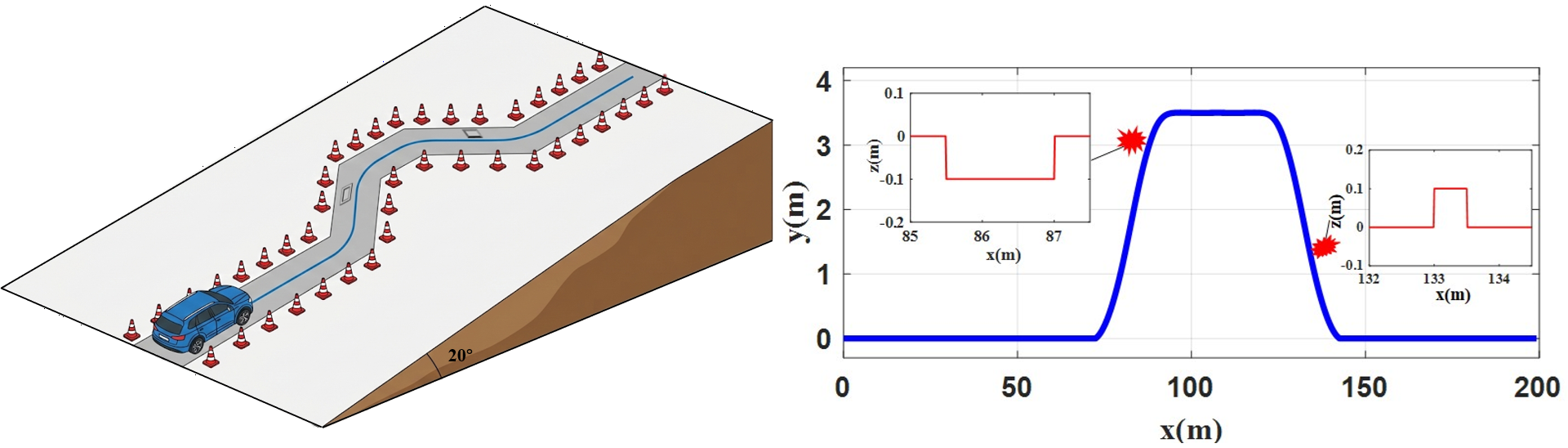}
	\caption{DLC on a coupled slope with potholes}
	\label{FIG:8}
\end{figure}

Double lane change (DLC) can directly reflect vehicle dynamic performance and has been widely used as a test case for path tracking. This paper designs a DLC scenario on a coupled slope with $\theta_s=20^{\circ}$, as shown in \figref{FIG:8}. In this scenario, a pothole with a depth of 0.1 m and a length of 1.5 m at $x=85.5 m$, and a bump with a height of 0.1 m and a length of 0.5 m at $x=133 m$, are included to validate the path tracking accuracy of the UGV on potholed road. The UGV longitudinal speed is set to 10 m/s.
\\
\textbf{Case 2: Complex off-road scenario}
\par
\begin{figure}
	\centering
		\includegraphics[width=0.95\textwidth]{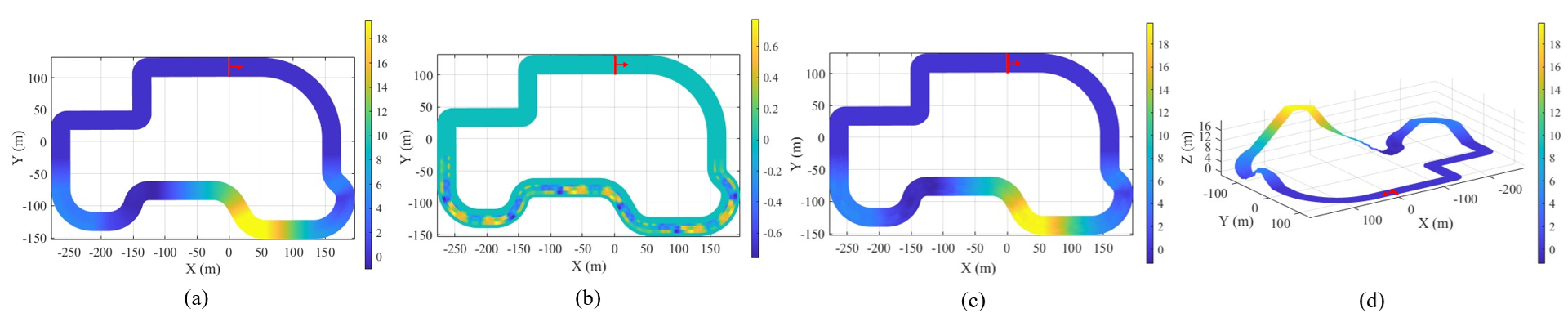}
	\caption{Complex off-road scenario}
	\label{FIG:9}
\end{figure}

To evaluate the path-tracking performance of CDK–LMPC on a realistic off-road scenario, a complex off-road scenario is designed as shown in \figref{FIG:9}. The colors in \figref{FIG:9}(a) and \figref{FIG:9}(b) represent the road elevation variations induced by the coupled slope and the potholed road, respectively. \figref{FIG:9}(c) shows the total elevation after combining the coupled slope and the potholed road. \figref{FIG:9}(d) presents the 3D schematic of the road. The UGV longitudinal speed is set to 7 m/s.

\subsection{Estimation of tire cornering stiffness}
\begin{figure}
	\centering
		\includegraphics[width=0.85\textwidth]{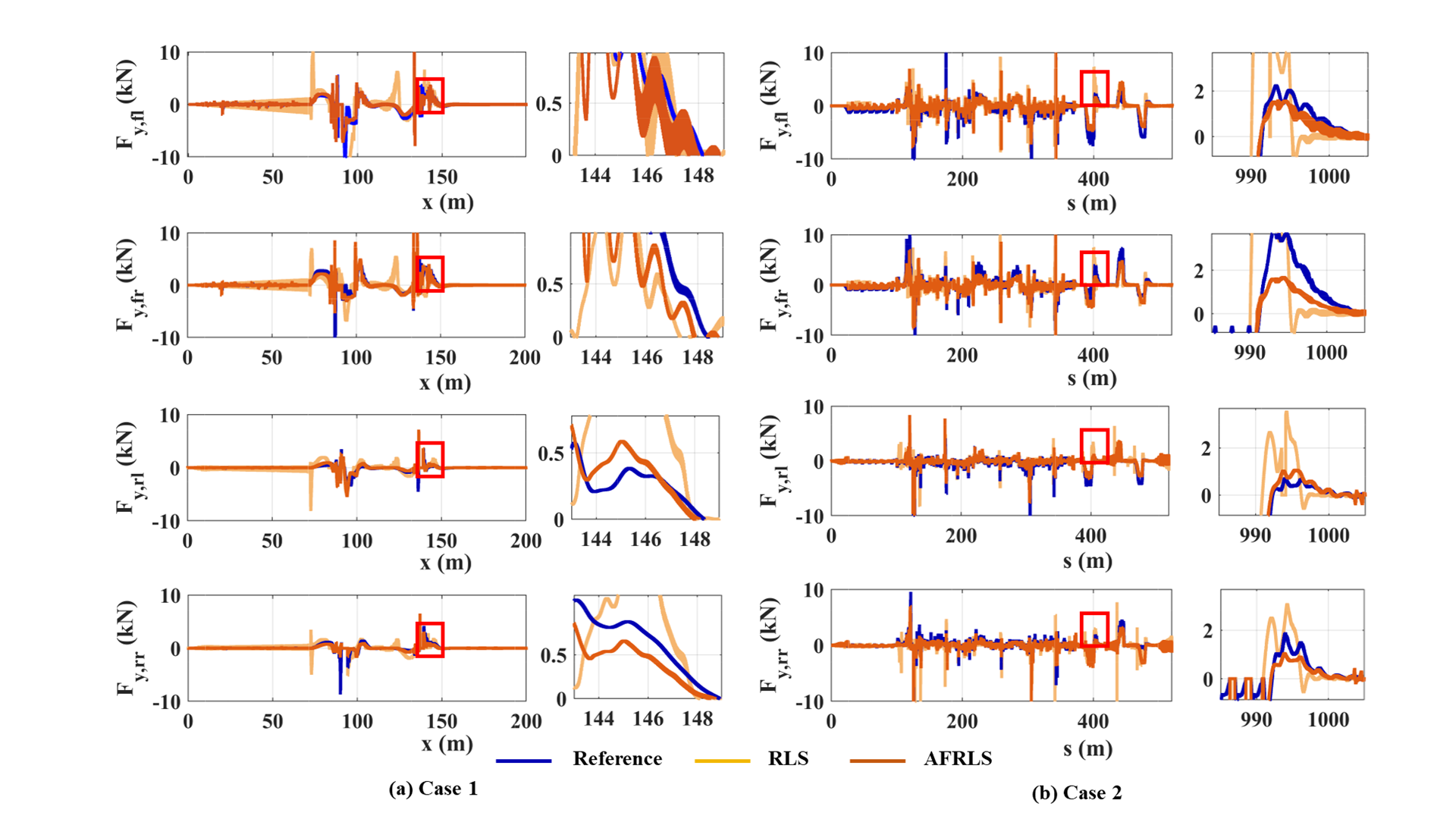}
	\caption{Estimation result of tire cornering force}
	\label{FIG:10}
\end{figure}

First, the effectiveness of the AFRLS-based tire cornering stiffness estimation method is validated under the above two scenarios. Since the experimental platform cannot directly output the tire cornering stiffness, the estimation accuracy of the four wheel lateral tire forces is compared, which can directly reflect the tire cornering stiffness, as shown in \figref{FIG:10}.
\par
Under Case 1, the RLS method exhibits large fluctuations in the estimates when the vehicle steers or traverses pothole surfaces, and thus cannot guarantee estimation accuracy. In contrast, the four wheel lateral forces estimated by the AFRLS method agree well with the ground truth values $F_{y,fl}, F_{y,rl}, F_{y,fr}$, and $F_{y,rr}$. In the absence of potholes, the AFRLS method can accurately capture the dynamic variations of the four wheel lateral forces, and the estimation error remains at a low level. When the vehicle passes over potholes, the four wheel lateral forces estimated by the AFRLS method exhibit brief fluctuations under transient road excitations, but quickly converge to the ground truth values after the vehicle leaves the potholes. Similarly, under Case 2, the AFRLS method still achieves high accuracy estimation of the four wheel lateral forces. This further demonstrates the accuracy and stability of the proposed tire cornering stiffness estimator under long duration severe disturbances, providing a solid foundation for precise UGV path tracking control.

\subsection{Experiment results and analysis}
Further, to demonstrate the superiority of the proposed CDK–LMPC, five comparative controllers are considered, whose components are summarized in TABLE~\ref{tab:3}:

\begin{table}
\caption{Components included in different control algorithms}
\label{tab:3}
\centering
\small
\renewcommand{\arraystretch}{1.15}
\begin{tabular}{lccc}
\toprule
\diagbox[width=10em]{Algorithms}{Components} & \makecell[c]{Laguerre\\function} & \makecell[c]{Event-Trigger} & \makecell[c]{Credibility\\coefficient} \\
\midrule
MPC        & $\times$ & $\times$ & $\times$ \\
LMPC       & $\checkmark$ & $\times$ & $\times$ \\
EDMD-LMPC  & $\checkmark$ & $\checkmark$ & $\times$ \\
KDMD-LMPC  & $\checkmark$ & $\checkmark$ & $\times$ \\
DK-LMPC    & $\checkmark$ & $\checkmark$ & $\times$ \\
CDK-LMPC   & $\checkmark$ & $\checkmark$ & $\checkmark$ \\
\bottomrule
\end{tabular}
\end{table}
MPC: Based on the state-space equations established in \secref{sec:2}, the MPC method is used for UGV path-tracking control.
\par
LMPC: The quadratic programming step in MPC is converted into a derivative based computation. The comparison between LMPC and MPC is used to illustrate the improvement brought by the Laguerre function to the control strategy.
\par
EDMD–LMPC: The lifting vector $\psi$ is constructed by manually selecting basis functions, and the finite-dimensional approximation matrix of the Koopman operator is then obtained to compensate for the path deviation under LMPC control. Specifically, $\psi$ is chosen as a combination of the auxiliary state $X$, 18 Gaussian radial basis functions, and 42 thin-plate spline radial basis functions. The specific functional form is:
\begin{equation}\label{(63)}
\varphi(x)_{\mathrm{Gauss}} = e^{-\Vert x - x_0 \Vert_2^2},
\end{equation}
\begin{equation}\label{(64)}
\varphi(x)_{\mathrm{Thinplate}} = \Vert x - x_0 \Vert_2^2 \cdot ln(\Vert x - x_0 \Vert_2),
\end{equation}
\par
KDMD–LMPC: The Koopman operator is approximated in a finite-dimensional form by implicitly constructing an infinite-dimensional feature space via the kernel trick, thereby compensating the UGV path deviation without manually constructing the lifting vector. The Gaussian radial basis function in \eqref{(63)} is selected as the kernel function.
\par
DK–LMPC: The lifting function $\psi$ is obtained by training the DNN designed in \secref{sec:4}, and the compensatory steering angle is predicted using the identified Koopman operator approximation matrix to compensate the path deviation under LMPC control.
\par
\begin{figure}
	\centering
		\includegraphics[width=0.9\textwidth]{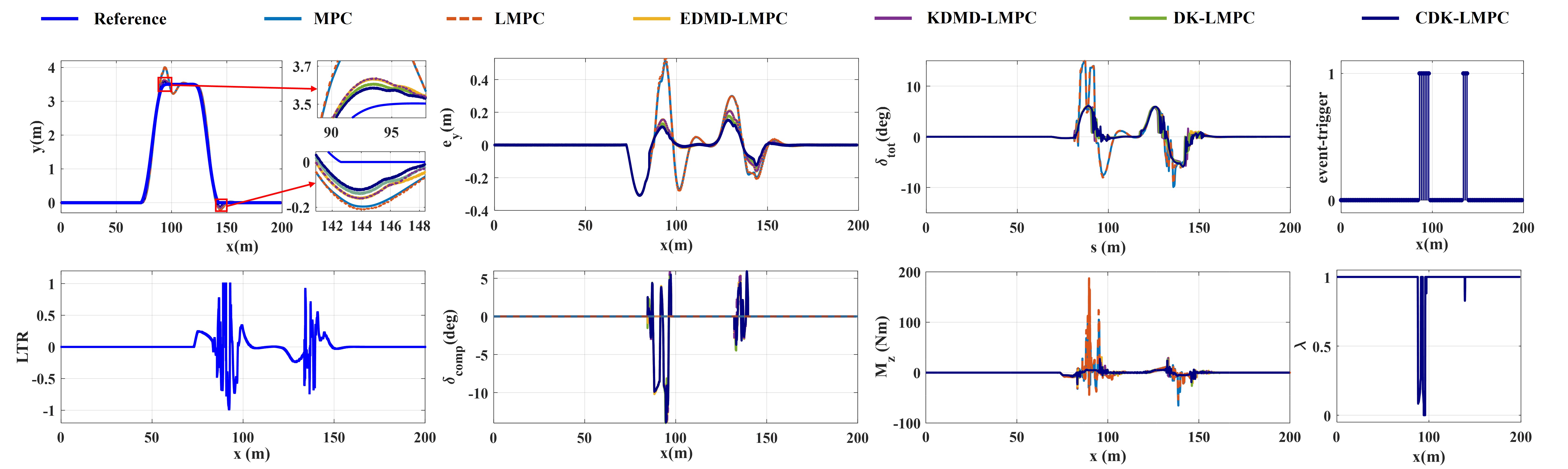}
	\caption{Performance of UGV path tracking under Case 1}
	\label{FIG:11}
\end{figure}
\begin{figure}
	\centering
		\includegraphics[width=0.9\textwidth]{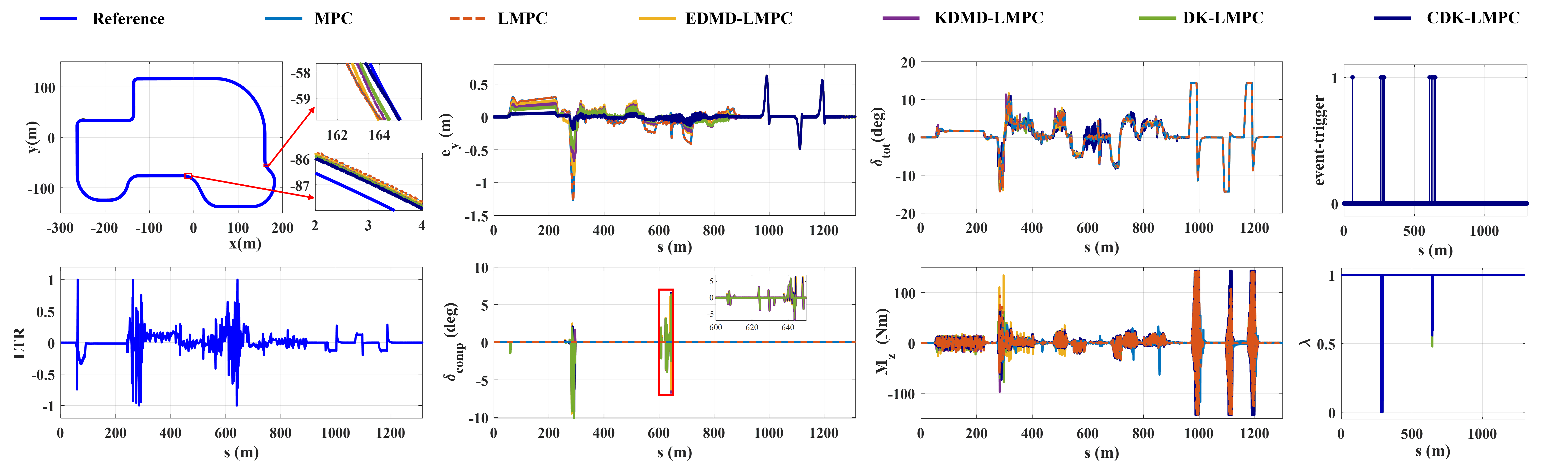}
	\caption{Performance of UGV path tracking under Case 2}
	\label{FIG:12}
\end{figure}

\figref{FIG:11} and \figref{FIG:12} show the path tracking performances of the UGV using the above control strategies under Case 1 and Case 2, respectively.
\par
For Case 1, the coupled slope and steering conditions alone have a limited effect on the LTR. However, under the influence of the potholed road, the LTR of UGV varies dramatically at $x=85-95 m$ and $x=133-140 m$. Accordingly, the EPC compensation mechanism can be effectively triggered in the potholed segments and generate an effective credibility coefficient $\lambda$ by selecting an appropriate $\mathrm{LTR}_\mathrm{thr}$. This ensures the execution feasibility of the overall steering command, thereby improving the path tracking stability of the UGV. In \figref{FIG:11}(c), the steering angles generated by MPC and LMPC exhibit obvious oscillations, causing the UGV to deviate significantly from the desired path. Moreover, the steering angle of LMPC is essentially consistent with that of MPC, resulting in highly coincident lateral errors $e_y$ for the two methods. When the LTR is large, EDMD–LMPC, KDMD–LMPC, DK–LMPC, and CDK–LMPC can significantly improve the path tracking accuracy of the UGV by generating the compensatory steering angle $\delta_{\mathrm{comp}}$. Similarly, for the more complex Case 2, when the LTR is large, EDMD–LMPC, KDMD–LMPC, DK–LMPC, and CDK–LMPC can also generate compensatory steering angles to improve path tracking accuracy.
\par
To quantitatively evaluate UGV path tracking performance, the root mean square lateral error $e_y^\mathrm{RMS}$ and the maximum lateral error $e_y^\mathrm{max}$ (compensated portion) of different control strategies under the above two cases are calculated, as shown in TABLE~\ref{tab:4}. It can be seen that the path tracking performances of UGVs using LMPC and MPC are similar, and the performance differences in both scenarios are less than 7\%. The proposed CDK–LMPC achieves the best $e_y^\mathrm{RMS}$ and $e_y^\mathrm{max}$ among the control strategies. It not only yields a substantial performance improvement compared with MPC, but also reduces $e_y^\mathrm{RMS}$ and $e_y^\mathrm{max}$ by 37.3\%, 59.7\%, 11.5\%, and 24.7\% relative to the advanced KDMD–LMPC in the two scenarios, respectively. This further demonstrates the superiority of the proposed CDK–LMPC. In addition, compared with CDK–LMPC, DK–LMPC produces compensatory steering angles with larger fluctuations and insufficient confidence, making it difficult to satisfy the physical constraints of the steering system. The superimposed overall steering angle also exhibits large oscillations, which affects tracking stability and consequently reduces the path tracking accuracy of the UGV.
\begin{table}
\caption{Path Tracking Results of Different Control Algorithm}
\label{tab:4}
\centering
\small
\renewcommand{\arraystretch}{1.15}
\begin{tabular}{c l c c}
\toprule
\textbf{Case} & \textbf{Performance indicators} & $\boldsymbol{e_y^{\max}}\,(\mathrm{m})$ & $\boldsymbol{e_y^{\mathrm{RMS}}}\,(\mathrm{m})$ \\
\midrule
\multirow{6}{*}{Case 1} 
& MPC        & 0.5138 & 0.1276 \\
& LMPC       & 0.5300 & 0.1304 \\
& EDMD-LMPC  & 0.2087 & 0.0793 \\
& KDMD-LMPC  & 0.2086 & 0.0789 \\
& DK-LMPC    & 0.1761 & 0.0734 \\
\rowcolor{gray!15}
& CDK-LMPC   & \textbf{0.1105} & \textbf{0.0698} \\
\midrule
\multirow{6}{*}{Case 2}
& MPC        & 1.2690 & 0.1886 \\
& LMPC       & 1.2935 & 0.2023 \\
& EDMD-LMPC  & 0.8779 & 0.1529 \\
& KDMD-LMPC  & 0.6593 & 0.1335 \\
& DK-LMPC    & 0.4761 & 0.1111 \\
\rowcolor{gray!15}
& CDK-LMPC   & \textbf{0.1918} & \textbf{0.0937} \\
\bottomrule
\end{tabular}
\end{table}
\par
\begin{figure}
	\centering
		\includegraphics[width=0.45\textwidth]{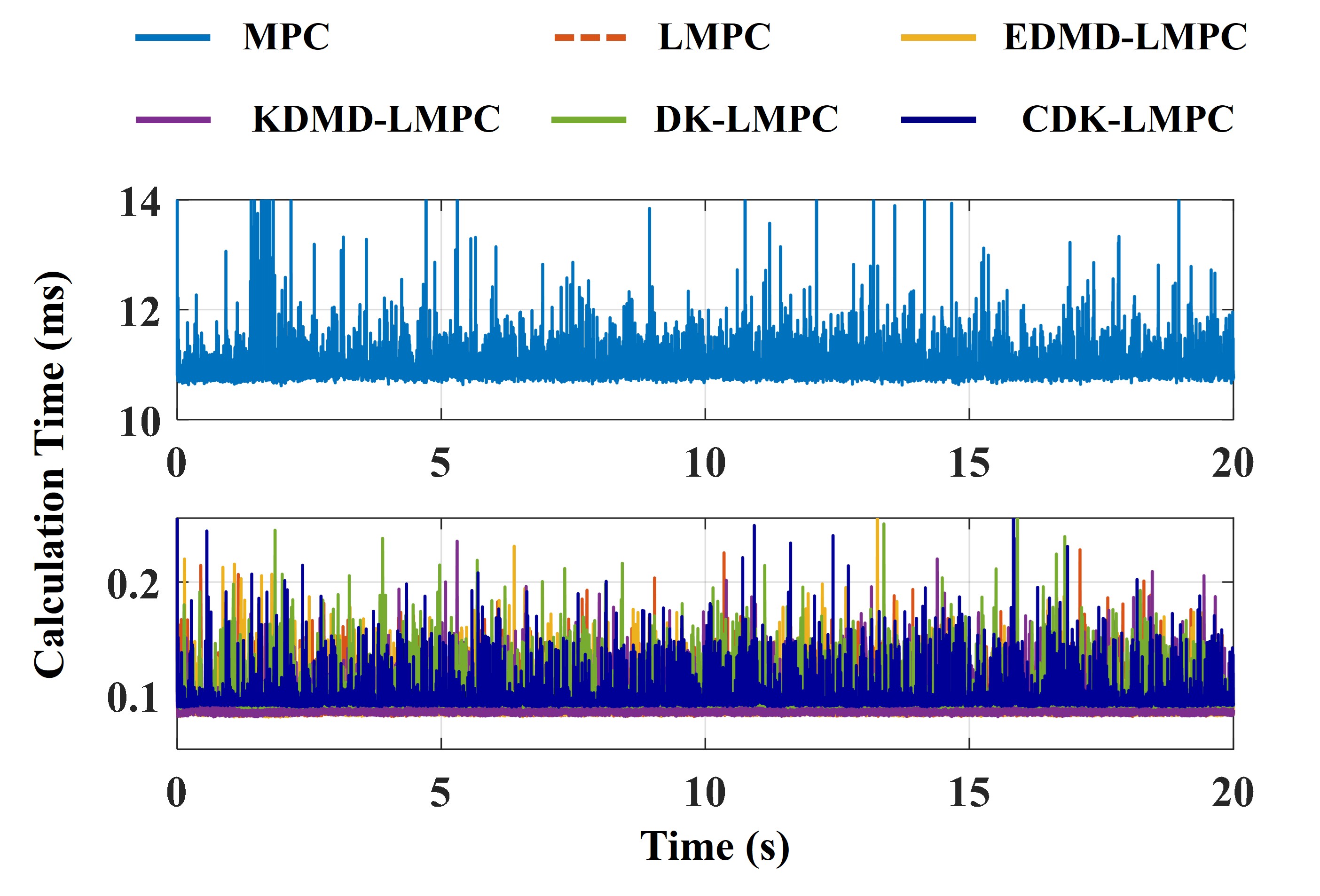}
	\caption{Actual computation time of different strategies}
	\label{FIG:13}
\end{figure}

To verify that LMPC reduces computational resource consumption compared with MPC, the computation time of each control strategy across different computational modules is compared, as shown in \figref{FIG:13}. The tests are conducted in MATLAB R2023a on an NVIDIA GeForce RTX 3060 and an Intel(R) Core(TM) i7-13700KF. The input sequence is solved using the quadprog quadratic programming solver in MPC, whereas the other control strategies use Laguerre functions for approximation. Furthermore, the maximum single step computation time $t_\mathrm{max}$ and the average single step computation time $t_\mathrm{mean}$ of different control strategies are calculated, as reported in TABLE~\ref{tab:5}. The statistical results indicate that LMPC-based control strategies can reduce the computation time by more than 98.89\% compared with MPC, significantly improving online efficiency and alleviating real-time computational burden.
\begin{table}
\caption{Calculation Time of Different Control Algorithm}
\label{tab:5}
\centering
\small
\renewcommand{\arraystretch}{1.15}
\begin{tabular}{p{0.4\textwidth} p{0.2\textwidth} p{0.2\textwidth}}
\toprule
\textbf{Performance indicators} & $t_{\max}\,(\mathrm{ms})$ & $t_{\mathrm{mean}}\,(\mathrm{ms})$ \\
\midrule
MPC        & 34.255 & 11.382 \\
LMPC       & 2.328  & \textbf{0.115} \\
EDMD-LMPC  & \textbf{1.842}  & 0.120 \\
KDMD-LMPC  & 1.981  & 0.119 \\
DK-LMPC    & 3.109  & 0.124 \\
\rowcolor{gray!15}
CDK-LMPC   & 5.461  & 0.126 \\
\bottomrule
\end{tabular}
\end{table}

\section{Conclusion}\label{sec:6}
This paper proposes a DNN Koopman-based deviation compensation strategy for UGV path tracking control. First, a path tracking state-space equation considering coupled slopes is established via coordinate transformation and dynamic analysis, and an LMPC path tracking controller is designed using Laguerre function. While reducing computational resource consumption, the proposed control strategy can ensure path tracking accuracy under different coupled slope conditions. Then, by integrating Koopman operator theory with DNN, a DK path deviation compensation method is developed, which significantly enhances the representational capability for the UGV nonlinear system on potholed road. Next, based on a path deviation compensation activation criterion and a credibility validation scheme, an event-triggered parallel cooperative compensation mechanism that integrates LMPC and DK is constructed, which improves the path tracking accuracy of the UGV on potholed road while ensuring the feasibility of the total steering angle after DK compensation. Finally, a hardware-in-the-loop experimental platform is built to validate the effectiveness of the proposed strategy. The experimental results show that the proposed AFRLS tire cornering stiffness estimation method can accurately characterize the variations of the four wheel cornering stiffness with potholed road excitations, achieving high estimation accuracy. The proposed CDK-LMPC strategy can stably constrain $e_y^\mathrm{max}$ and $e_y^\mathrm{RMS}$ within 0.2 m and 0.1 m in complex off-road scenarios, respectively. Compared with the MPC method, CDK-LMPC improves the UGV path tracking performance by more than 45.3\% while reducing the computation time by approximately 98.89\%. This substantially enhances operational efficiency in addition to the significant improvement in path tracking performance.

\section*{Declaration of competing interest}
The authors declare that they have no known competing financial interests or personal relationships that could have appeared to influence the work reported in this paper.

% Use the style of numbering in square brackets.
% If nothing is used, default style will be taken.
%\begin{enumerate}[a)]
%\item 
%\item 
%\item 
%\end{enumerate}  

%\begin{itemize}
%\item 
%\item 
%\item 
%\end{itemize}  

%\begin{description}
%\item[]
%\item[] 
%\item[] 
%\end{description}  

% Uncomment and use as the case may be
%\begin{theorem} 
%\end{theorem}

% Uncomment and use as the case may be
%\begin{lemma} 
%\end{lemma}


\printcredits

\bibliographystyle{elsarticle-num}

\bibliography{cas-refs}



\end{document}